\documentclass[journal]{IEEEtran}
\ifCLASSINFOpdf
\else
\fi
\usepackage{amssymb}
\usepackage{hyperref}
\usepackage{amsmath,bm}
\usepackage{algorithm}
\usepackage{algpseudocode}
\usepackage{cite}
\usepackage{epstopdf}
\usepackage{subfig}
\usepackage{tabularx}
\usepackage{tabulary}
\usepackage{mathrsfs}
\usepackage{graphicx}
\usepackage{multirow}
\usepackage{makecell}
\usepackage{booktabs}
\usepackage{flushend}
\begin{document}
\title{Restricted Minimum Error Entropy Criterion\\ for Robust Classification}
\author{Yuanhao~Li,
        Badong~Chen,~\IEEEmembership{Senior~Member,~IEEE,}
        Natsue~Yoshimura,
        and~Yasuharu~Koike
\thanks{This work was supported in part by JST PRESTO (Precursory Research for Embryonic Science and Technology) under Grant JPMJPR17JA, and was also supported by National Natural Science Foundation of China (91648208, 61976175). \emph{(Corresponding author: Yuanhao Li.)}}
\thanks{Y. Li, N. Yoshimura, and Y. Koike are with Institute of Innovative Research, Tokyo Institute of Technology, Yokohama, Japan (e-mail: li.y.ay@m.titech.ac.jp; yoshimura@pi.titech.ac.jp; koike@pi.titech.ac.jp).}
\thanks{B. Chen is with Institute of Artificial Intelligence and Robotics, Xi'an Jiaotong University, Xi'an, China (e-mail: chenbd@mail.xjtu.edu.cn).}
\thanks{N. Yoshimura is also with PRESTO, Japan Science and Technology Agency, Japan.}}
\maketitle
\begin{abstract}
The \emph{minimum error entropy} (MEE) criterion has been verified as a powerful approach for non-Gaussian signal processing and robust machine learning. However, the implementation of MEE on robust classification is rather a vacancy in the literature. The original MEE only focuses on minimizing the Renyi's quadratic entropy of the error probability distribution function (PDF), which could cause failure in noisy classification tasks. To this end, we analyze the optimal error distribution in the presence of outliers for those classifiers with continuous errors, and introduce a simple codebook to restrict MEE so that it drives the error PDF towards the desired case. Half-quadratic based optimization and convergence analysis of the new learning criterion, called \emph{restricted MEE} (RMEE), are provided. Experimental results with logistic regression and extreme learning machine are presented to verify the desirable robustness of RMEE.
\end{abstract}


\section{Introduction}
\label{sec1}
\IEEEPARstart{M}{any} tasks in machine learning require robustness — that the learning process of a model is less affected by noises than by regular samples \cite{hampel1986robust}. Different from the noise in regression which means that attribute value diverges from the foreseeable distribution, the noise in classification is more intractable and can be systematically classified into two categories: attribute noise and label noise \cite{zhu2004class,quinlan1986induction}. The attribute (or feature) noise means measurement errors resulting from noisy sensors, recordings, communications, and data storage, while the label noise means a mistake when labeling samples. As stated in \cite{frenay2013classification}, label noise could sometimes result from mutual elements as attribute noise, such as communication errors, whereas it mainly arises from expert elements \cite{hickey1996noise}: i) unreliable labeling due to insufficient information, ii) unreliable non-expert for low cost, and iii) subjective labeling. Not to mention, classes are not always totally distinguishable as \emph{lived} and \emph{died} \cite{bross1954misclassification}. The outlier, a more severe case of noise \cite{collett1976subjective}, usually causes serious performance degradation. According to the above taxonomy, we state that attribute outliers signify deviate attribute values but completely irrelevant to label information, and label outliers imply that some distinct samples are assigned with wrong labels. Note that mislabeled samples are not necessarily label outliers since they could occur near the boundary region thus being less adverse for learning machine \cite{frenay2013classification}.

Consider binary classification here. The discriminant function $f(X)$ is learned from the given training samples $\{(x_i, t_i) \}_{i=1}^{N}$ by empirical risk minimization of a loss function $\mathcal{L}(T,f(X))$, where $x_i \in \mathbb{R}^d$ is the attribute value of the \emph{i}th sample and $t_i \in \{-1,1\}$ is the label. In this paper, we use uppercase letter to represent a random variable, and lowercase letter to represent its value. Generally, loss functions are designed with respect to the margin $z=tf(x)$, written as $\mathcal{L}(z)$. The minimization of 0-1 loss $\mathcal{L}_{0-1}(z)=\lVert \max (0,-z) \rVert _0$ leads to minimum misclassification rate on training dataset directly, whereas its optimization is intractable. Therefore, many alternatives were proposed by using convex upper bounds of $\mathcal{L}_{0-1}(z)$ \cite{zhang2004statistical,bartlett2006convexity}. For example, the hinge loss $\mathcal{L}_{hinge}(z)=\max(0,1-z)$ is used in the support vector machine (SVM), the exponential loss $\mathcal{L}_{exp}(z)=\exp(-z)$ is used in the AdaBoost, and the logistic regression applies the logistic loss $\mathcal{L}_{log}(z)=\log(1+\exp(-z))$.

However, it is shown that the above classifiers based on convex loss functions are not robust to outliers \cite{zhu2004class,frenay2013classification}. This mainly arises from the unbounded property of the convex loss functions, which would assign large losses on outliers \cite{frenay2013classification,wu2007robust,masnadi2009design,miao2015rboost}. Consequently, the learning process is mainly determined by outliers, rather than those meaningful samples, and the decision boundaries could be affected severely, leading to significant performance degradation.

For robust classification, many algorithms have been proposed to suppress the adverse effects of outliers. One intuitive approach is to remove or relabel training samples in data preprocessing \cite{frenay2013classification,hodge2004survey,chandola2009anomaly,feng2014robust}, whereas this could possibly ignore useful information in training dataset. Weighting samples is another widely used method which aims to reduce the outliers' proportion in the learning process \cite{frenay2013classification,suykens2002weighted,byrnes2018kernel}. In addition, the recovery of clean data by robust principal component analysis can realize robust classification as well \cite{yin2018robust}. Moreover, meta-learning technique can achieve robustness by evaluating gradients for each data point at the learned parameters \cite{dia2019sever}.

To achieve robust classification, it is an alternative way to make the learning process itself robust, which means using a bounded loss function so that it will not assign large values for outliers thus being robust. In \cite{masnadi2009design}, the bounded Savage loss was proposed to construct the robust SavageBoost algorithm, and \cite{miao2015rboost} further extended this work. In \cite{ertekin2010nonconvex}, one robust SVM algorithm was developed based on the ramp loss. In \cite{yang2014robust}, the truncated least square loss was proposed for the robust least square SVM. Simulation results in the above works have shown the effectiveness of using a bounded and non-convex loss function for robust classification.

The \emph{information theoretic learning} (ITL) has been proved to be promising for robust machine learning. The \emph{minimum error entropy} (MEE) criterion is one fundamental and popular approach in this field, which usually aims to minimize the quadratic Renyi's entropy of training errors. MEE has been utilized to propose state-of-the-art robust algorithms for regression \cite{principe2010information,chen2019quantized}, feature extraction \cite{chen2018common}, dimensionality reduction \cite{he2010principal,guo2004modified}, subspace clustering \cite{wang2015minimum,li2019robust}, and so on. By contrast, the potential robustness of MEE with respect to outliers in classification has not been thoroughly explored. In this paper, we aim to propose an implementation of MEE for robust classification. 

The remainder of this paper is organized as follows. In Section \ref{sec2}, it is expounded how to treat classification from the perspective of error rather than margin. In Section \ref{sec3}, we analyze the optimal error distribution in the presence of outliers. In Section \ref{sec4}, we give a brief introduction of the original MEE and its quantized version, and interprets its potential failure in classification tasks. In Section \ref{sec5}, based on the optimal error distribution with outliers, we design a specific codebook to restrict MEE, proposing the restricted MEE criterion. In Section \ref{sec6}, experimental results on logistic regression and extreme learning machine for toy datasets and benchmark datasets, respectively, are presented. Next, we provide some discussions in Section \ref{sec7}. Finally, Section \ref{sec8} gives the conclusion.

\section{View Classification from Error}
\label{sec2}
Consider logistic regression here, which is one of the most widely used models. The logistic loss $\mathcal{L}_{log}(z)=\log(1+\exp(-z))$ can be interpreted from a more principle perspective, cross entropy. In what follows, label $T \in \{0,1\}$ is usually used because this leads to a simpler form of logistic regression. With a fixed parameter $\omega \in \mathbb{R}^d$, the probability that $x_i$ belongs to class 1 is predicted as
\begin{equation}
y_i=P(t_i=1)=\frac{1}{1+\exp(-\omega' x_i)}
\end{equation}
in which $\omega'$ is transpose of $\omega$ and the mapping from $\omega' x_i \in (-\infty, \infty)$ to probability $y_i \in (0, 1)$ is the well known \emph{sigmoid}. Based on the assumed Bernoulli distribution, the opposite probability for class 0 is $P(t_i=0)=1-y_i$. The parameter $\omega$ can be learned by maximizing the cross entropy (CE) between the true label $T$ and predicted probability $Y$ based on the Kullback-Leibler divergence, which leads to the following empirical risk minimization
\begin{equation}
\begin{split}
\omega^*&=\mathop{arg\min}_{\omega \in \mathcal{W}} \hat{\mathcal{R}}_{CE}(Y)\; \\  \label{lrce}
&=\mathop{arg\min}_{\omega \in \mathcal{W}} -\sum_{i=1}^{N}{((1-t_i)\log(1-y_i)+t_i\log(y_i))}\; \\
&=\mathop{arg\min}_{\omega \in \mathcal{W}} -\sum_{t_i=0}{\log(1-y_i)}-\sum_{t_i=1}{\log(y_i)}\; \\
\end{split}
\end{equation}
in which $\mathcal{W}$ stands for the parameter space. This form is actually equivalent to minimizing the logistic loss $\mathcal{L}_{log}(z)$ in the context of $T \in \{-1,1\}$. 

The purpose of (\ref{lrce}) is to maximize the similarity between $T$ and $Y$, which can be regarded as minimizing the difference as well. The error $E=T-Y$ is one basic variable to describe the difference between two variables. Although (\ref{lrce}) does not contain $e_i$ explicitly, \cite{de2013minimum} gives the following derivation.

\noindent \textbf{Derivation 1\label{deri1}}: To use $\{-1,1\}$-label scheme instead of $\{0,1\}$-label scheme, one is supposed to use the \emph{tanh} transformation to obtain the prediction, or to convert the prediction through $y\rightarrow(y+1)/2$ from the \emph{sigmoid} transformation. Thus in the context of $\{-1,1\}$-label scheme, the empirical risk $\hat{\mathcal{R}}_{CE}(Y)$ in (\ref{lrce}) is rewritten as
\begin{equation}
\hat{\mathcal{R}}_{CE}(Y)=-\sum_{t_i=-1}{\log(\frac{1-y_i}{2})}-\sum_{t_i=1}{\log(\frac{1+y_i}{2})}
\end{equation}
which is the empirical version of the following theoretical risk
\begin{equation}
\begin{split}
\mathcal{R}_{CE}(Y)=&-P(-1)\int{\log(1-y)f_{Y\mid -1}(y)dy}\; \\  \label{theorisk}
&-P(1)\int{\log(1+y)f_{Y\mid 1}(y)dy}+\log 2\; \\
\end{split}
\end{equation}
where $P(\cdot)$ is class prior probability and $f_{Y\mid \cdot}(y)$ is class-conditional probability density function (PDF) evaluated at $y$. Substituting $E=T-Y$ into (\ref{theorisk}) and ignoring the constant $\log 2$, one can obtain
\begin{equation}
\mathcal{R}_{CE}(E)=\sum_{t\in \{-1,1\}}{-P(t)\int{\log(2-te)f_{E\mid t}(e)de}}
\end{equation}

Thus, one obtains the loss function w.r.t. error $\mathcal{L}_{CE}(t,e)=-\log (2-te)$. $\hfill\blacksquare$

\begin{figure}[t!]
	\centering
	\includegraphics[width=0.8\columnwidth]{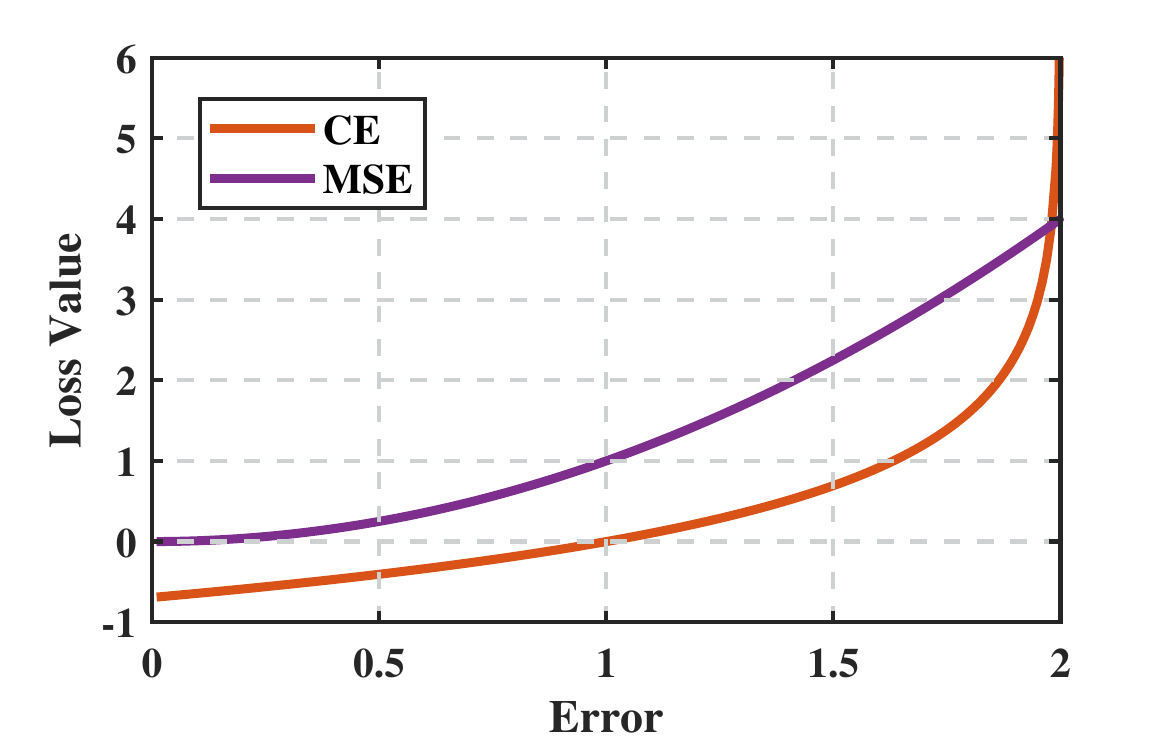}
	\caption{Loss curves of $\mathcal{L}_{CE}(t,e)$ and $\mathcal{L}_{MSE}(e)$ ($t=1$).}
	\label{loss_curves}
\end{figure}

Note that in learning process, the actual predicted value is not the discrete label but the probability $Y$ of continuous value, which means one can obtain continuous errors. In $\{-1,1\}$-label scheme, where $T \in \{-1,1\}$ and $Y \in (-1,1)$, one obtains $E = T - Y$ that belongs to the continuous open interval $(-2,2)$ by subtraction. Therefore, it is also feasible to learn by minimizing the \emph{mean squared error} (MSE). However, it will lead to non-convexity if MSE loss $\mathcal{L}_{MSE}(e)=e^2$ is used with nonlinear transformation \emph{tanh}. Nevertheless, using MSE with \emph{tanh} brings benefits instead. In Fig. \ref{loss_curves}, we illustrate the loss curves of $\mathcal{L}_{CE}(t,e)$ and $\mathcal{L}_{MSE}(e)$ in the interval $e \in (0,2)$ when $t=1$. One can see that when $e$ is close to its maximum $e\rightarrow 2$, $\mathcal{L}_{CE}(t,e)$ will approach infinity, which means CE is unbounded and non-robust. By contrast, $\mathcal{L}_{MSE}(e)$ is always no more than $4$ because one has $|e|<2$. Thus $\mathcal{L}_{MSE}(e)$ is bounded, which means MSE could be robust potentially if used with \emph{tanh}. Note that the above arguments hold for \emph{sigmoid} as well, since \emph{sigmoid} could be obtained from \emph{tanh} with simple translation and scaling. That is to say, in context of $\{0,1\}$-label scheme, in which $T \in \{0,1\}$ and $Y \in (0,1)$, one can obtain continuous error $E\in (-1,1)$ by subtraction.

Moreover, currently in a variety of neural networks for classification, \emph{tanh} and \emph{sigmoid} transformation is widely used in output layers. According to \cite{de2013minimum}, these classifiers with continuous errors, such as logistic regression and neural networks, are named as regression-like classifiers. On the contrary, those with discrete errors are named as non-regression-like classifiers, such as decision trees where the prediction is discrete label rather than continuous probability. 

\section{Error Distribution Analysis}
\label{sec3}
In this section, in the context of $\{0,1\}$-label coding scheme and logistic regression, we focus on analyzing the optimal error distribution $\rho_E(e)$ in the presence of outliers. Denoting the class prior probability by $p=P(T=1)$ and $q=1-p=P(T=0)$, one obtains the cumulative distribution function of error as follows
\begin{equation}
\begin{split}
F_{E}(e)&=P(E \leqslant e)\; \\  
&=pP(E \leqslant e|T=1)+qP(E \leqslant e|T=0)\; \\
&=pP(1-Y \leqslant e|T=1)+qP(-Y \leqslant e|T=0)\; \\
&=p(1-F_{Y|1}(1-e))+q(1-F_{Y|0}(-e))\; \\
&=1-pF_{Y|1}(1-e)-qF_{Y|0}(-e)\; \\
\end{split}
\end{equation}
The error PDF is obtained by differentiation of $F_{E}(e)$ as
\begin{equation}
f_E(e)=pf_{Y|1}(1-e)+qf_{Y|0}(-e)   \label{FEE}
\end{equation}
Assume that the samples of 0 class and 1 class are of two multivariate Gaussian distributions, $X|_{T=0}\sim\mathcal{N}(x;\mu _0,\varSigma _0)$ and $X|_{T=1}\sim\mathcal{N}(x;\mu _1,\varSigma _1)$, respectively. Given the boundary parameter $\omega$, then $\omega'X|_T$ is of a univariate Gaussian distribution $\mathcal{N}(\omega'x;\omega'\mu _{T},\omega'\varSigma _{T} \omega)$. To obtain the PDF of the univariate random variable $Y|_T=1/(1+\exp(-\omega'X|_T))$, we give the following well-known theorem.

\begin{figure}[t!]
	\centering
	\includegraphics[width=0.75\columnwidth]{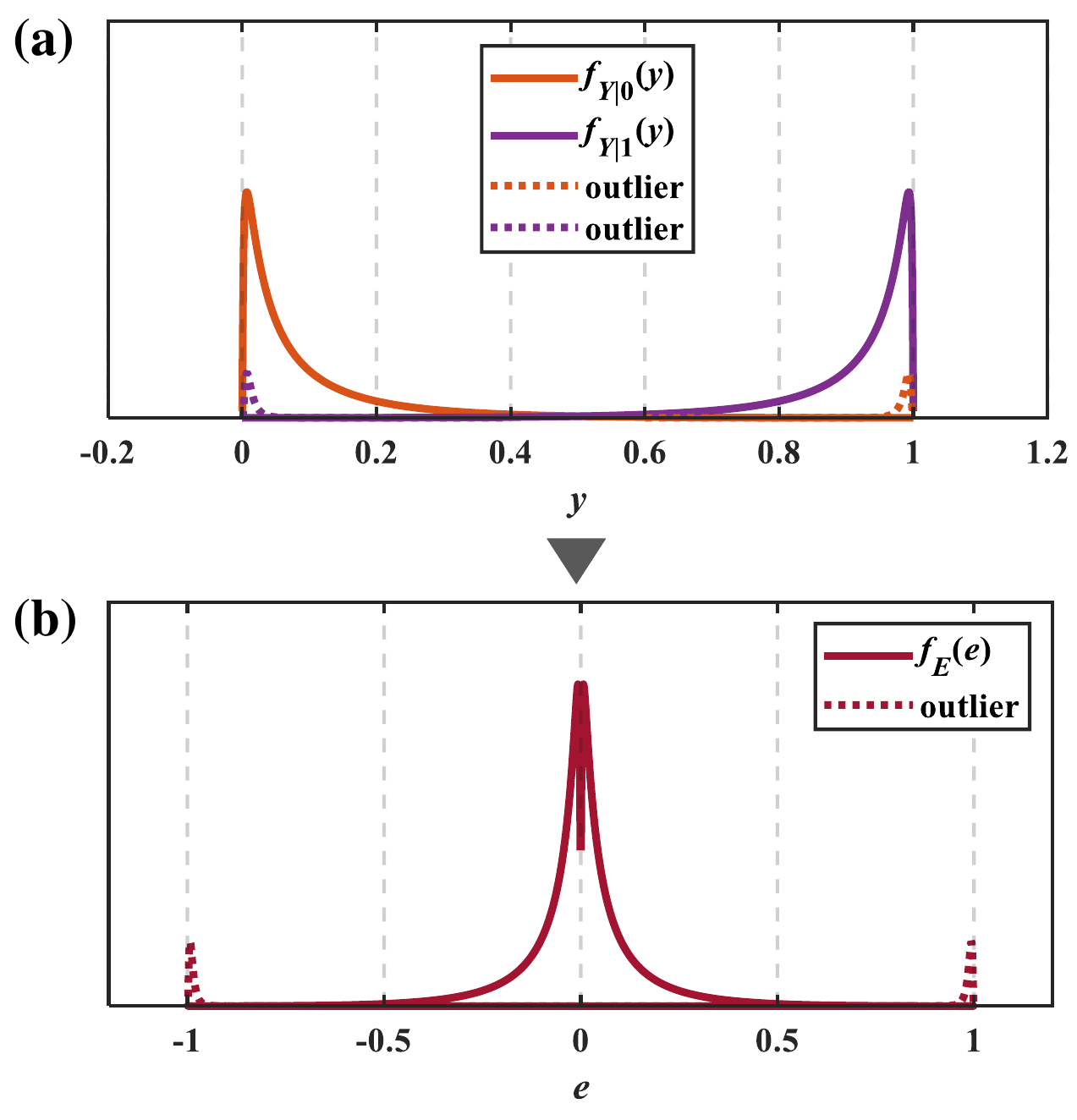}
	\caption{Illustrative function curves of (a)$f_{Y|T}(y)$ and (b)$f_E(e)$ ($\omega'\mu _{0}=-5$, $\omega'\mu _{1}=5$, $\omega'\varSigma _{0}\omega=\omega'\varSigma _{1}\omega=5$, $p=q=0.5$).}
	\label{fig_error_dist}
\end{figure}

\noindent \textbf{Theorem 2\label{theo2}}: Assume $f_X(x)$ is the PDF of a random variable $X$, and $\vartheta(x)$ is a monotonic and differentiable function. If $g_Y(y)$ is the PDF of $Y=\vartheta(X)$ and $\vartheta'(x)\ne0$, $\forall x\in X$, then 
\begin{equation}
g_Y(y)=\begin{cases}
\frac{f_X(\vartheta^{-1}(y))}{|\vartheta'(\vartheta^{-1}(y))|}\,\,\,\, \text{inf}\vartheta(x)<y<\text{sup}\vartheta(x) \\
0\,\,\,\,\,\,\,\,\,\,\,\,\,\,\,\,\,\,\,\,\,\,\,\,\,\,\,\,\, \text{otherwise} \\
\end{cases}
\end{equation}
where $x=\vartheta^{-1}(y)$ is the inverse function of $y=\vartheta(x)$.$\hfill\blacksquare$
Since the \emph{sigmoid} function satisfies the above conditions, one can obtain the PDF of $Y|_T$ as
\begin{equation}
f_{Y|T}(y) = \frac{\exp(-\frac{(\log(\frac{y}{1-y})-\omega'\mu _{T})^2}{2\omega'\varSigma _{T} \omega})}{\sqrt{2\pi\omega'\varSigma _{T} \omega}}  \label{FYT}
\end{equation}
where $0< y< 1$. This PDF can be viewed as a nonlinear scaling on the horizontal axis of Gaussian distribution, and hence it is single-peak as well. Since the optimal parameter aims to achieve minimum misclassification rate, it is supposed that most of the predicted $Y|_T$ are as close to the corresponding $T$ as possible, to say the distribution peak of $f_{Y|1}(y)$ and $f_{Y|0}(y)$ to be close to $1$ and $0$, respectively. Intuitive function curves with specific $\omega'\mu _{T}$ and $\omega'\varSigma _{T}\omega$ are given in Fig. \ref{fig_error_dist}(a) with solid lines. If adverse outliers happen to $X|_T$, the corresponding prediction will approach the opposite, since this is what \emph{outlier} means. That is to say, for example, $f_{Y|1}(y)$ will emerge a small peak near $0$, and vice versa for $f_{Y|0}(y)$. The distributions caused by outliers are illustrated in dashed lines in Fig. \ref{fig_error_dist}(a). Substituting (\ref{FYT}) into (\ref{FEE}), one can obtain 
\begin{equation}
\begin{split}
f_E(e)=&\frac{p}{\sqrt{2\pi\omega'\varSigma _{1} \omega}}\exp(-\frac{(\log(\frac{1-e}{e})-\omega'\mu _{1})^2}{2\omega'\varSigma _{1} \omega})\; \\ 
+&\frac{q}{\sqrt{2\pi\omega'\varSigma _{0} \omega}}\exp(-\frac{(\log(\frac{-e}{1+e})-\omega'\mu _{0})^2}{2\omega'\varSigma _{0} \omega})\; \\  
\end{split}
\end{equation}
which is plotted with solid lines in Fig. \ref{fig_error_dist}(b) with the same specific $\omega'\mu _{T}$ and $\omega'\varSigma _{T}\omega$ as above, and the effects of outliers are shown with dashed lines as well. One can observe three significant peaks on $\{0,-1,1\}$. 

Not only Gaussian cases but also others usually lead to a similar $f_E(e)$. Even in the cases where $X|_T$ is of multi-peak distribution, i.e. $\omega'X|_T$ is multi-peak, $f_E(e)$ could be probably similar. The reason is, $\omega'X|_T$ distributed on $(-\infty, \infty)$ is squeezed to $Y|_T\in (0, 1)$ by \emph{sigmoid}, and thus multiple peaks could be close enough so that they can be viewed as one peak. Moreover, the above arguments hold for other regression-like classifiers as well, since as stated before, the predicted $Y|_T$ is supposed to be distributed close to the corresponding $T$ in any kind of regression-like classifiers.

Now we focus on giving the formalization of the desired three-peak distribution $\rho_E(e)$. For clarity, supposing that class $0$ stands for negative and class $1$ means positive, we denote those outliers that lie in the positive side but are assigned with negative labels as false negative (\emph{FN}) outliers, and vice versa as false positive (\emph{FP}) outliers. Note that $x_i$ is predicted correctly if the corresponding $|e_i|<0.5$, otherwise wrongly. Therefore, the errors brought by inliers would be less than $0.5$ in the sense of absolute value since they are supposed to be classified correctly. On the other hand, \emph{FN} outliers result in errors belonging to $(-1, -0.5)$, and \emph{FP} outliers lead to errors belonging to $(0.5, 1)$. For simplicity, we assume that each peak is close enough to a Dirac-$\delta$ function so that the density of the desired error PDF is zero beyond the three peaks. As a result, $\rho_E(e)$  is denoted as
\begin{equation} \label{dist}
\rho_E(e)=\begin{cases}
\zeta_0\,\,\,\,\,\,\,\,\,\,\,\,\,\,\,\, e=0 \\
\zeta_{-1}\,\,\,\,\,\,\,\,\,\,\,\, e=-1 \\
\zeta_{1}\,\,\,\,\,\,\,\,\,\,\,\,\,\,\,\, e=1 \\
0\,\,\,\,\,\,\,\,\,\,\,\,\,\,\,\,\,\,\, \text{otherwise} \\
\end{cases}
\end{equation}
where $\zeta_i$ ($i=0,-1,1$) denotes the corresponding density for each peak. One may have noticed that $\zeta_i$ is closely related to the proportion of each type of samples. To be specific, $\zeta_0$ is the proportion of inliers since the corresponding peak results from those samples that are supposed to be classified correctly. Similarly, $\zeta_1$ (or $\zeta_{-1}$) is the proportion of \emph{FP} (or \emph{FN}) outliers.

If, as in cross entropy loss $\mathcal{L}_{CE}(t,e)$, a large penalty is imposed to a large error, outliers will be dominant in the learning process, thus making it difficult to learn through meaningful samples. The inspiration of this study is that, the optimal parameter will result in an error distribution that holds three significant peaks at $\{0,-1,1\}$ by inliers, \emph{FN} outliers, and \emph{FP} outliers, respectively. If a classifier is designed to realize a similar error distribution, it can probably achieve satisfactory robust classification.

\section{MEE for Classification}
\label{sec4}
\subsection{Introduction of MEE Criterion}
The \emph{minimum error entropy} (MEE) criterion has been proved to be robust in many machine learning tasks. MEE aims to minimize the Renyi's $\alpha$ entropy of prediction error, which is introduced as a generalization of Shannon's entropy \cite{principe2010information}. Renyi's $\alpha$ entropy, or called Renyi's entropy of $\alpha$-order, is defined as 
\begin{equation}
{{H}_{R,\alpha }}\left( f_E(e) \right) \triangleq \frac{1}{1-\alpha }\log \int{f_{E}^{\alpha}\left(e\right) de}
\end{equation}
in which $f_E(e)$ denotes the PDF of prediction error. The \emph{information potential} is defined as the term in the logarithm
\begin{equation}
{I_{\alpha}}\left( f_E(e) \right) \triangleq \int{f_{E}^{\alpha}\left( e \right) de=E}\left[ f_{E}^{\alpha -1}\left( e \right) \right]  \label{ip}
\end{equation}
where $E[\cdot]$ is the expectation operator. For simplicity, the parameter $\alpha$ is usually set at $\alpha=2$. Because the logarithm is monotonically increasing, minimizing Renyi's quadratic entropy ${{H}_{R,2}}\left( f_E(e) \right)$ is equal to maximizing the quadratic information potential ${I_{2}}\left( f_E(e) \right)$
\begin{equation}
\begin{split}
\min H_{R,2}\left( f_E(e) \right) &\Longleftrightarrow \max I_2\left( f_E(e) \right) \; \\   \label{equ}
&\Longleftrightarrow \max E\left[ f_E(e) \right]\; \\
\end{split}
\end{equation}
Based on an empirical version of the quadratic information potential \cite{principe2010information}, one can obtain
\begin{equation}
\begin{split}
\omega^*&=\mathop{arg\max}_{\omega \in \mathcal{W}} \hat{I}_{2}\left( f_E(e) \right)\; \\ \label{mee}
&=\mathop{arg\max}_{\omega \in \mathcal{W}} E\left[ \hat{f}_E(e)\right]=\mathop{arg\max}_{\omega \in \mathcal{W}}\frac{1}{N}\sum_{i=1}^N{\hat{f}_E\left( e_i \right)} \; \\
&=\mathop{arg\max}_{\omega \in \mathcal{W}}\frac{1}{N^2}\sum_{i=1}^N{\sum_{j=1}^N{\kappa _{\sigma}\left( e_i-e_j \right)}} \; \\
\end{split}
\end{equation}
where $\hat{f}_E\left( \cdot \right) $ is the estimated error PDF by Parzen's estimator \cite{silverman1986density,parzen1962estimation}
\begin{equation}
\hat{f}_E\left( e \right) =\frac{1}{N}\sum_{j=1}^N{\kappa _{\sigma}\left( e-e_j \right)}
\end{equation}
and $\kappa _{\sigma}\left( \cdot \right) $ is the Gaussian kernel function with bandwidth $\sigma$
\begin{equation}
{{\kappa }_{\sigma }}\left( x \right)=\frac{1}{\sqrt{2\pi }\sigma }\exp \left( -\frac{{{x}^{2}}}{2{{\sigma }^{2}}} \right) \label{kerf}
\end{equation}
One could view the PDF estimator $\hat{f}_E\left( \cdot \right) $ as an adaptive objective function since it changes with $\{e_i\}_{i=1}^{N}$, which is different from the conventional ones that are generally invariable. The adaptation is advantageous, which has been proved theoretically as well as confirmed numerically \cite{principe2010information}. 

To alleviate the computational bottleneck caused by double summation in (\ref{mee}), quantization technique is implemented that the error PDF is estimated by a little part of samples but not the entirety, thus decreasing the number of inner summation \cite{chen2019quantized}. As a result, the quantized MEE (QMEE) is expressed as
\begin{equation}
\begin{split}
\omega^*&=\mathop{arg\max}_{\omega \in \mathcal{W}} \hat{I}_{2}^Q\left( f_E(e) \right)\; \\ \label{qmee}
&=\mathop{arg\max}_{\omega \in \mathcal{W}} E\left[ \hat{f}_{E}^{Q}(e)\right]=\mathop{arg\max}_{\omega \in \mathcal{W}}\frac{1}{N}\sum_{i=1}^N{\hat{f}_{E}^{Q}\left( e_i \right)}\; \\
&=\mathop{arg\max}_{\omega \in \mathcal{W}}\frac{1}{N^2}\sum_{i=1}^N{\sum_{j=1}^N{\kappa _{\sigma}\left( e_i-Q[e_j] \right)}}\; \\
&=\mathop{arg\max}_{\omega \in \mathcal{W}} \frac{1}{N^2}\sum_{i=1}^N{\sum_{j=1}^M{\varphi_j\kappa _{\sigma}\left( e_i-c_j \right)}} \; \\
\end{split}
\end{equation}
where $\hat{I}_{2}^Q\left( f_E(e) \right)$ denotes the quantized quadratic information potential and $\hat{f}_{E}^{Q}(e)$ is the estimated error PDF based on some representative samples. $Q[\cdot]$ denotes a quantization operator that leads to a codebook $C=(c_1,c_2,...,c_M)$, which means $Q[\cdot]$ is a function that maps each $\{e_i\}_{i=1}^{N}$ to one of $\{c_j\}_{j=1}^{M}$. The parameter $\varPhi=(\varphi_1,\varphi_2,...,\varphi_M)$ denotes the number that how many samples are quantized to the corresponding word. Clearly one has $\sum_{j=1}^M{\varphi_j}=N$. Since $\{c_j\}_{j=1}^{M}$ is a representative description of $\{e_i\}_{i=1}^{N}$, we usually have $M\ll N$ and the complexity is thus decreased from $O(N^2)$ to $O(MN)$. Proved by theoretical analysis and experimental results, QMEE can realize commensurate performance as the original one with proper quantization \cite{chen2019quantized,chen2018common}. In short, it is precisely because the codebook words $\{c_j\}_{j=1}^{M}$ are representative enough for the entirety $\{e_i\}_{i=1}^{N}$, that QMEE in (\ref{qmee}) can play the same effect as the original MEE in (\ref{mee}). In the context of univariate error, an adaptive quantization method was proposed in \cite{chen2018common}, which is summarized in Algorithm \ref{algo_quan}.
 
\begin{algorithm}[t]
	\caption{Procedure of Adaptive Quantization}\label{algo_quan}
	\begin{algorithmic}[1]
		\State Input samples $\left\{ {{x}_{i}} \right\}_{\text{i}=1}^{N}$. Compute the range $L=\max(x_i)-\min(x_i)$.
		\State Parameter setting: quantization threshold $\varepsilon$. Usually $\varepsilon=0.05$.
		\State Initialize ${{C}_{1}}=\left\{ {{x}_{1}} \right\}$, where ${{C}_{i}}$ denotes the codebook at the \emph{i}th iteration.
		\For{$i=2,\cdots ,N$}
		\State Compute the minimum distance between ${{x}_{i}}$ and ${{C}_{i-1}}$:
		$\text{dis}\left( {{x}_{i}},{{C}_{i-1}} \right)=\underset{1\le j\le \left| {{C}_{i-1}} \right|}{\mathop{\min}}\left| {{x}_{i}}-{{C}_{i-1}}\left( {{j}} \right) \right|$
		where ${{C}_{i-1}}\left( j \right)$ denotes the \emph{j}th element of ${{C}_{i-1}}$, and $\left| {{C}_{i-1}} \right|$ stands for the size of ${{C}_{i-1}}$.
		\If{$\text{dis}\left( {{x}_{i}},{{C}_{i-1}} \right)\le \varepsilon L $}
		\State Keep the codebook unchanged: ${{C}_{i}}={{C}_{i-1}}$ and quantize ${{x}_{i}}$ to the closest code word: $Q\left[ {{x}_{i}} \right]={{C}_{i-1}}\left( {{j}^{*}} \right)$, where ${j}^{*}=\underset{1\le j\le \left| {{C}_{i-1}} \right|}{\mathop{\arg\min}}\left| {{x}_{i}}-{{C}_{i-1}}\left( {{j}} \right) \right|$.
		\Else
		\State Update the codebook: ${{C}_{i}}=\left\{ {{C}_{i-1}},{{x}_{i}} \right\}$ and quantize ${{x}_{i}}$ to itself: $Q\left[ {{x}_{i}} \right]={{x}_{i}}$.
		\EndIf
		\EndFor
		\State Output $\left\{ Q\left[ {{x}_{i}} \right] \right\}_{i=1}^{N}$.
	\end{algorithmic}
\end{algorithm}

Entropy provides a PDF concentration measure that higher concentration implies lower entropy, which is the initial motivation to use entropic risk functionals. For continuous distributions, the local minimum value of ${{H}_{R,2}}\left( f_E(e) \right)$ corresponds to a PDF represented by several continuous Dirac-$\delta$ functions, a Dirac-$\delta$ comb. When all errors are zero, a single Dirac-$\delta$ at the origin for the error PDF can be achieved, leading to the ideal situation, $f_E(e)=0|_{e\ne0}$. This demands a learning machine, in iterative training, to indeed guarantee the convergence of the error PDF towards a single Dirac-$\delta$ at the origin.

The robustness of MEE can be briefly explained as follows. In the training process with regular samples, MEE ensures most of errors are close to zero so as to approach a Dirac-$\delta$ function at the origin. If outliers happen, the error PDF will not only hold a main peak at the origin as before, but also generate small peaks at large errors caused by outliers. This kind of distribution, as mentioned earlier, is a local minimum for MEE as well. In addition, it can be interpreted from the perspective of equation (\ref{mee}) and (\ref{kerf}). For a large error caused by outlier, its effect on the maximized term $\hat{I}_2\left( f_E(e) \right)$ in (\ref{mee}) is weakened since the Gaussian kernel function in (\ref{kerf}) is bounded, which can saturate the summation term $\kappa _{\sigma}\left( e_i-e_j \right)$ of the last line in (\ref{mee}). Theoretical insights of robustness are provided in \cite{principe2010information,chen2016insights}.

\subsection{MEE for Classification}
Through the brief introduction, MEE is supposed to be appropriate for robust classification, since the optimal error PDF with outliers is a three-peak distribution, which is exactly an optimal for MEE. Nevertheless, compared to regression, one will encounter additional suffering in the implementation of MEE for classification. In the literature, one valuable study explored the implementation of MEE for different classifiers \cite{de2013minimum}. As stated, ``MEE is harder for classification than for regression''. The main difficulties are summarized as follows.

In binary classification, according to \cite{de2013minimum}, the purpose of MEE can be decomposed as 
\begin{equation}
\begin{split}
&\min H_{R,2}\left( f_E(e) \right) \; \\  
&\Longleftrightarrow \max I_2\left( f_E(e) \right)\; \\
&\Longleftrightarrow\max (p^2I_2\left( f_{Y|1}(1-e) \right) + q^2I_2\left( f_{Y|0}(-e) \right)) \; \\  
\end{split}
\end{equation}
in which the class-conditional property causes the difficulty. Recall that class-conditional distributions, entropies, and information potentials depend on the model parameter $\omega$, although this dependency has been omitted for simpler notation. Minimizing $H_{R,2}\left( f_E(e) \right)$ implies maximizing the sum of $p^2I_2\left( f_{Y|1}(1-e) \right)$ and $q^2I_2\left( f_{Y|0}(-e) \right)$, both of which are functions with respect to $\omega$. Thus, it is difficult to say about the minimum of $H_{R,2}\left( f_E(e) \right)$ since it depends on $p$, $q$, $f_{Y|1}$, and $f_{Y|0}$ simultaneously. One has to consider each class-conditional distribution individually and study them together with the weights $p$, $q$ to achieve minor $H_{R,2}\left( f_E(e) \right)$ as possible. By contrast, in regression tasks, $f_E(e)$ is not divided into several class-conditional parts but as a whole, which is much easier to deal with.

The above interpretation seems not intuitive that how MEE could fail for classification, so we provide a specific scenario. Sometimes MEE based classifiers may predict all samples as the same class with large confidence. For example, suppose that each predicted probability $\{y_i\}_{i=1}^{N}$ is close to $0$. Thus, the errors from 0-class samples will be close to $0$, while those from 1-class samples will be close to $1$, resulting in an $f_E(e)$ with two approximate Dirac-$\delta$ functions at $\{0, 1\}$, respectively. The basic explanation was already given as before: any Dirac-$\delta$ comb achieves local minimum entropy. Note that when a similar case occurs, the classification accuracy could be even the chance level. This instability of MEE for classification is in particular explained in \cite{de2013minimum} with illustrative examples, which is verified in this paper as well by experimental results.

\section{Restricted MEE}
\label{sec5}
The instability above inspires that, only focusing on minimizing entropy, i.e. maximizing information potential in (\ref{mee}), is not enough. From now on, getting rid of the MEE framework temporarily, we first focus on driving the error PDF obtained by the training process towards the optimal three-peak distribution $\rho_E(e)$ in (\ref{dist}). To make two distributions as similar as possible, a basic idea is to maximize a similarity measure between their PDFs. A quantity of similarity measures for PDF exist in the literature, for which \cite{cha2007comprehensive} provides a comprehensive survey. In this paper we utilize the fundamental \emph{inner product} to measure the similarity between distributions, which is generalized from its use for vectors \cite{deza2006dictionary,duda2012pattern}. The inner-product similarity between two continuous PDFs $f_X(x)$ and $g_X(x)$ is computed as
\begin{equation}
\left \langle f_X(x),g_X(x)\right \rangle =\int_X{f_X(x)g_X(x)dx}
\end{equation}
Now one can maximize this similarity measure between the error PDF $f_E(e)$ and the desired distribution $\rho_E(e)$ as
\begin{equation}
\begin{split}
&\max \left \langle f_E(e),\rho_E(e)\right \rangle  \; \\   \label{rmee1}
& \Longleftrightarrow \max \int_X{f_E(e)\rho_E(e)dx} \; \\
&\Longleftrightarrow \max \zeta_0f_E(0)+\zeta_{-1}f_E(-1)+\zeta_{1}f_E(1)\; \\
\end{split}
\end{equation}
the last equality of which arises from that $\rho_E(e)$ in (\ref{dist}) is always zero except when $e= 0,-1,$ or $1$. In practice, one maximizes the similarity using the empirical version as
\begin{equation}
\begin{split}
\omega^*&=\mathop{arg\max}_{\omega \in \mathcal{W}} \left \langle \hat{f}_E(e),\rho_E(e)\right \rangle     \; \\   \label{rmee2}
&=\mathop{arg\max}_{\omega \in \mathcal{W}}\zeta_0\hat{f}_E(0)+\zeta_{-1}\hat{f}_E(-1)+\zeta_{1}\hat{f}_E(1)     \; \\
&=\mathop{arg\max}_{\omega \in \mathcal{W}} {\left( \begin{array}{c}
	\zeta_0\frac{1}{N}\sum_{i=1}^N{\kappa _{\sigma}\left( 0-e_i \right)} \\
	+\zeta_{-1}\frac{1}{N}\sum_{i=1}^N{\kappa _{\sigma}\left( -1-e_i \right)} \\
	+\zeta_1\frac{1}{N}\sum_{i=1}^N{\kappa _{\sigma}\left( 1-e_i \right)}  \\
	\end{array} \right)} \; \\
&=\mathop{arg\max}_{\omega \in \mathcal{W}} \frac{1}{N}\sum_{i=1}^N{\left( \begin{array}{c}
	\zeta_0\kappa _{\sigma}\left( e_i \right) \\
	+\zeta_{-1}\kappa _{\sigma}\left( e_i+1 \right) \\
	+\zeta_1\kappa _{\sigma}\left( e_i-1 \right)  \\
	\end{array} \right)} \; \\
&=\mathop{arg\max}_{\omega \in \mathcal{W}} \frac{1}{N^2}\sum_{i=1}^N{\left( \begin{array}{c}
	N\zeta_0\kappa _{\sigma}\left( e_i \right) \\
	+N\zeta_{-1}\kappa _{\sigma}\left( e_i+1 \right) \\
	+N\zeta_1\kappa _{\sigma}\left( e_i-1 \right)  \\
	\end{array} \right)} \; \\
\end{split}
\end{equation}
One may have noticed the comparability between this form and QMEE, since this formula (\ref{rmee2}) can be regarded as a special case of QMEE in (\ref{qmee}) when the codebook $C=(0,-1,1)$, the corresponding quantization number $\varPhi=(N\zeta_0,N\zeta_{-1},N\zeta_{1})$, and obviously $M=3$. Note that the derivation of formula (\ref{rmee2}) has nothing to do with the MEE framework originally since it aims to maximize the inner-product similarity between the error PDF $f_E(e)$ and the desired distribution $\rho_E(e)$. That is to say, from another point of view, we obtain a consequence that are closely related to MEE.

Now, returning back to MEE framework, we will interpret the meaning of formula (\ref{rmee2}). From the perspective of principle, QMEE aims to concentrate the prediction errors as close as possible to each $\{c_j\}_{j=1}^{M}$ to achieve a relatively narrow error distribution, in which $\{\varphi_j\}_{j=1}^{M}$ act as weight parameters. We can expect that if the codebook is assigned with some specific values, QMEE will focus the training errors close to these positions with certain weights $\varPhi$. With this consideration, we implement QMEE with a predetermined codebook $C=(0,-1,1)$, the purpose of which is to restrict errors on these three positions, avoiding the undesirable double-peak training consequence. QMEE with a restricted codebook, \emph{restricted MEE} (RMEE) in short, is proposed by using the predetermined codebook $C=(0,-1,1)$, which is denoted as 
\begin{equation} 
\begin{split}
\omega^*&=\mathop{arg\max}_{\omega \in \mathcal{W}} \hat{I}_{2}^R\left( f_E(e) \right) \; \\ \label{rmee3}
&=\mathop{arg\max}_{\omega \in \mathcal{W}} \frac{1}{N^2}\sum_{i=1}^N{\left( \begin{array}{c}
	\varphi_0\kappa _{\sigma}\left( e_i \right) \\
	+\varphi_{-1}\kappa _{\sigma}\left( e_i+1 \right) \\
	+\varphi_1\kappa _{\sigma}\left( e_i-1 \right)  \\
	\end{array} \right)}\; \\
\end{split}
\end{equation}
where $\hat{I}_{2}^R\left( f_E(e) \right)$ denotes the restricted quadratic information potential and we have $\varPhi=(\varphi_0,\varphi_{-1},\varphi_{1})=(N\zeta_0,N\zeta_{-1},N\zeta_{1})$, which denotes the corresponding number for each quantization word $C=(0,-1,1)$. One can see obviously that the essential difference between QMEE and the proposed RMEE is, the codebook $C$ of the former is obtained by a data-driven method as in Algorithm \ref{algo_quan}, which aims to make the elements $\{c_j\}_{j=1}^{M}$ as representative to the entirety as possible, while the latter's is predetermined, which aims to drive the error PDF $f_E(e)$ towards the desired one $\rho_E(e)$.

Now, through the above interpretations, two perspectives for the proposed RMEE are summarized as follows. 

\noindent \textbf{1}: RMEE can be regarded as maximizing the inner-product similarity between the error PDF $f_E(e)$ and the desired three-peak distribution $\rho_E(e)$. 

\noindent \textbf{2}: RMEE can also be viewed as a special case of QMEE that the codebook is predetermined as $C=(0,-1,1)$, which aims to concentrate errors on these three locations as possible.

In prediction (or testing), the method of labeling a sample is the same as in traditional methods that use the \emph{sigmoid} transformation. Given the resultant model parameter, one first computes the probability $y_i$ for each sample according to the model. For example, in logistic regression, one has $y_i=1/(1+\exp(-\omega' x_i))$. Then, one is supposed to label the sample $x_i$ with class 1 if the corresponding $y_i>0.5$, otherwise with class 0.

In the following, for the proposed RMEE, we discuss about its optimization, convergence analysis, and how to determine the hyper-parameters.

\subsection{Optimization}
As seen in (\ref{rmee3}), the Gaussian kernel function will bring non-convexity in optimization, not to mention the implicit \emph{sigmoid} transformation which is intractable particularly. Here we utilize the \emph{half-quadratic} (HQ) technique to solve this problem, which is often used to solve ITL optimization issues \cite{yuan2009robust,he2011robust,xu2016robust,ren2018correntropy}. To derive the HQ-based optimization for (\ref{rmee3}), we first give the following theorem.

\noindent \textbf{Theorem 3}:\label{theo3} Define a convex function $g(v)=-v\log(-v)+v$, where $v<0$. Based on the conjugate function theory \cite{boyd2004convex}, one has
\begin{equation}
\exp(-\frac{(t-y)^2}{2\sigma^2})=\mathop{\sup}_{v<0}\{v\frac{(t-y)^2}{2\sigma^2}-g(v)\}      \label{hq_ori}
\end{equation}
where the supremum is achieved at $v=-\exp(-\frac{(t-y)^2}{2\sigma^2})<0$. See the proof in \cite{xu2016robust,ren2018correntropy}.$\hfill\blacksquare$

\noindent Thus the form of RMEE in (\ref{rmee3}) can be rewritten as
\begin{equation} 
\begin{split}
\label{rmee_hq_1}
&\omega^*=\mathop{arg\max}_{\omega \in \mathcal{W}} \sum_{i=1}^N{\left( \begin{array}{c}
	\varphi_0\mathop{\sup}_{u_i<0}\{u_i\frac{e_{i}^2}{2\sigma^2}-g(u_i)\} \\
	+\varphi_{-1}\mathop{\sup}_{v_i<0}\{v_i\frac{(e_{i}+1)^2}{2\sigma^2}-g(v_i)\} \\    
	+\varphi_1\mathop{\sup}_{s_i<0}\{s_i\frac{(e_{i}-1)^2}{2\sigma^2}-g(s_i)\}  \\     
	\end{array} \right)}\; \\
&=\mathop{arg\max}_{\omega \in \mathcal{W},u_i<0,v_i<0,s_i<0} \sum_{i=1}^N{\left( \begin{array}{c}
	\varphi_0(u_i\frac{e_{i}^2}{2\sigma^2}-g(u_i)) \\
	+\varphi_{-1}(v_i\frac{(e_{i}+1)^2}{2\sigma^2}-g(v_i)) \\    
	+\varphi_1(s_i\frac{(e_{i}-1)^2}{2\sigma^2}-g(s_i))  \\     
	\end{array} \right)}\; \\
&=\mathop{arg\max}_{\omega \in \mathcal{W},u_i<0,v_i<0,s_i<0}  J_R(\omega,u_i,v_i,s_i)\; \\
\end{split}
\end{equation}
where $1/(N^2\sqrt{2\pi}\sigma)$ is omitted in the second equality since it is a constant with fixing the kernel bandwidth $\sigma$. Note that although the model parameter $\omega$ is implicit in $J_R(\omega,u_i,v_i,s_i)$ in (\ref{rmee_hq_1}), it has a direct influence on $e_i$. Now one can optimize $J_R(\omega,u_i,v_i,s_i)$ by alternate optimization on $\omega$, $u_i$, $v_i$, and $s_i$, respectively. To be specific, in the \emph{k}th iteration with the current errors $\{e_i \}_{i=1}^{N}$, one first optimizes 
\begin{equation} 
\begin{split}  \label{hq_op_1}
&(u_i^k,v_i^k,s_i^k)\; \\
&=\mathop{arg\max}_{u_i<0,v_i<0,s_i<0} \sum_{i=1}^N{\left( \begin{array}{c}  
	\varphi_0(u_i\frac{e_{i}^2}{2\sigma^2}-g(u_i)) \\
	+\varphi_{-1}(v_i\frac{(e_{i}+1)^2}{2\sigma^2}-g(v_i)) \\    
	+\varphi_1(s_i\frac{(e_{i}-1)^2}{2\sigma^2}-g(s_i))  \\     
	\end{array} \right)}\; \\
&=\mathop{arg\max}_{u_i<0,v_i<0,s_i<0}  J_{R1}(u_i,v_i,s_i)\; \\
\end{split}
\end{equation}
According to (\ref{hq_ori}), the closed-form solution of (\ref{hq_op_1}) is 
\begin{equation} 
\label{hq_op_2}
\begin{array}{c}       
u_i^k=-\exp(-\frac{e_{i}^2}{2\sigma^2})<0 \\
v_i^k=-\exp(-\frac{(e_{i}+1)^2}{2\sigma^2})<0 \\    
s_i^k=-\exp(-\frac{(e_{i}-1)^2}{2\sigma^2})<0 \\     
(i=1,2,...,N) \\     
\end{array}
\end{equation}
Second, after obtaining the optimal $(u_i^k,v_i^k,s_i^k)$ in the \emph{k}th iteration, one obtains $\omega^k$ by solving the following optimization
\begin{equation} 
\begin{split} \label{hq_op_3}
\omega^k&=\mathop{arg\max}_{\omega \in \mathcal{W}} \sum_{i=1}^N{\left( \begin{array}{c}
	\varphi_0(u_i\frac{e_{i}^2}{2\sigma^2}-g(u_i)) \\
	+\varphi_{-1}(v_i\frac{(e_{i}+1)^2}{2\sigma^2}-g(v_i)) \\    
	+\varphi_1(s_i\frac{(e_{i}-1)^2}{2\sigma^2}-g(s_i))  \\     
	\end{array} \right)}\; \\
&=\mathop{arg\max}_{\omega \in \mathcal{W}} \sum_{i=1}^N{\left( \begin{array}{c}
	\varphi_0u_i(t_i-y_i)^2 \\
	+\varphi_{-1}v_i(t_i+1-y_i)^2   \\    
	+\varphi_1s_i(t_i-1-y_i)^2    \\     
	\end{array} \right)}\; \\
&=\mathop{arg\max}_{\omega \in \mathcal{W}} J_{R2}(\omega)\; \\
\end{split}
\end{equation}
in which $g(\cdot)$ and $1/2\sigma^2$ are omitted since they are constants in this step. Note that for different regression-like classifiers, the form of $y_i$ is different. For example, in logistic regression one has $y_i=1/(1+\exp(-\omega' x_i))$, while $y_i$ will be sophisticated in neural networks since they are of hierarchical structures. Nevertheless, even in neural networks, one can always optimize $J_{R2}(\omega)$ in (\ref{hq_op_3}) with the prominent back-propagation technique \cite{hecht1992theory} and gradient-based optimization, since the objective function $J_{R2}(\omega)$ is continuous and differentiable. Here we give the derivation in the context of logistic regression, in which the gradient of $J_{R2}(\omega)$ in (\ref{hq_op_3}) is
\begin{equation} 
\begin{split} \label{grad}
&\frac{\partial J_{R2}(\omega)}{\partial \omega}=\sum_{i=1}^N{\left( \begin{array}{c}
	\varphi_0u_i\frac{\partial (t_i-y_i)^2}{\partial \omega} \\
	+\varphi_{-1}v_i\frac{\partial (t_i+1-y_i)^2}{\partial \omega}   \\      
	+\varphi_1s_i\frac{\partial (t_i-1-y_i)^2}{\partial \omega}    \\        
	\end{array} \right)}\; \\
&= -2\sum_{i=1}^N{\left( \begin{array}{c}
	\varphi_0u_ie_i \\
	+\varphi_{-1}v_i(e_i+1)   \\      
	+\varphi_1s_i(e_i-1)    \\        
	\end{array} \right)x_iy_i(1-y_i)}\; \\
\end{split}
\end{equation}
Then one can use gradient-based or momentum-based optimization, such as the popular and efficient \emph{Adam} \cite{kingma2014adam}, to obtain $\omega^k$ for (\ref{hq_op_3}). The HQ-based optimization for RMEE is summarized in Algorithm \ref{hqrmee}. Note that in (\ref{hq_op_1})(\ref{hq_op_2})(\ref{hq_op_3})(\ref{grad}), we omit the superscript \emph{k} somewhere for the reason of clarity.

\begin{algorithm}[!htb]
	\caption{HQ-based Optimization for RMEE}
	\label{hqrmee}
	\begin{algorithmic}[1]
		\State \textbf{Input}:
		training samples $\{x_i, t_i \}_{i=1}^{N}$;
		initial model parameter $\omega$; 
		hyper-parameters $\sigma$, $\varPhi=(\varphi_0,\varphi_{-1},\varphi_{1})$;
		a small positive value $\varsigma$
		\State \textbf{Output}:
		model parameter $\omega$
		\Repeat 
		\State Initialize $converged=$ FALSE; 
		\State Compute the errors $\{e_i \}_{i=1}^{N}$ at the current model parameter $\omega$; 
		\State Update $(u_i,v_i,s_i)$ with (\ref{hq_op_2}); 
		\State Update $\omega$ with (\ref{hq_op_3}) by gradient-based or momentum-based optimization;
        \If{ the difference of the objective function in (\ref{rmee3}) is smaller than $\varsigma$}
		\State $converged=$ TRUE
		\EndIf
		\Until $converged==$ TRUE
	\end{algorithmic}
\end{algorithm}

\subsection{Convergence Analysis}
One may be concerned about whether Algorithm \ref{hqrmee} can converge to a local maximum. Actually, convergence of HQ-based optimization can be easily proved as
\begin{equation} 
\begin{split} \label{hq_con1}
J_R(\omega^{k-1},u_i^{k-1},v_i^{k-1},s_i^{k-1})&\leqslant J_R(\omega^{k-1},u_i^{k},v_i^{k},s_i^{k})\; \\
&\leqslant J_R(\omega^{k},u_i^{k},v_i^{k},s_i^{k})\; \\
\end{split}
\end{equation}
in which the first inequality is established obviously according to (\ref{hq_op_1})(\ref{hq_op_2}). To establish the second inequality, i.e. $J_R(\omega^{k-1},u_i^{k},v_i^{k},s_i^{k})\leqslant J_R(\omega^{k},u_i^{k},v_i^{k},s_i^{k})$, the following equivalent inequality is supposed to be established with fixing $(u_i,v_i,s_i)=(u_i^{k},v_i^{k},s_i^{k})$
\begin{equation} 
\label{hq_con2}
J_{R2}(\omega^{k-1})\leqslant J_{R2}(\omega^{k})
\end{equation}
Thus, to guarantee the convergence of Algorithm \ref{hqrmee}, one could find it not necessary for $\omega$ to achieve maximum in (\ref{hq_op_3}). On the contrary, as long as we have $J_{R2}(\omega^{k-1})\leqslant J_{R2}(\omega^{k})$ at every iteration with fixing $(u_i,v_i,s_i)=(u_i^{k},v_i^{k},s_i^{k})$, the inequality (\ref{hq_con1}) is established, which guarantees the convergence of Algorithm \ref{hqrmee}. Therefore, for the optimization problem in (\ref{hq_op_3}), one only needs to consider whether the new $\omega^{k}$ achieves a larger objective function value than the previous $\omega^{k-1}$, i.e. $J_{R2}(\omega^{k-1})\leqslant J_{R2}(\omega^{k})$.

\subsection{Hyper-Parameters Determination}
\label{hyper}
Now, we focus on solving the determination of hyper-parameters $\sigma$ and $\varPhi$ for the proposed RMEE.

The kernel bandwidth $\sigma$ plays a vital role in Parzen-window-based methods. The famous \emph{Silverman's Rule} was proposed in \cite{silverman1986density} for density estimation, which has been used in ITL methods \cite{he2011robust}. However, this method is not always favorable for ITL methods \cite{principe2010information,chen2019quantized,chen2018common}. Therefore, in this paper we use the conservative five-fold cross-validation to choose a proper $\sigma$ for RMEE.

Considering the determination of $\varPhi$, the optimal values are supposed to be the numbers of inliers, \emph{FN} outliers, and \emph{FP} outliers, corresponding to $\varphi_0$, $\varphi_{-1}$, and $\varphi_1$, respectively. However, this will be intractable unless we have prior information about the outlier proportion. To determine $\varPhi$ without any prior information, we utilize the following empirical method to obtain an approximate estimation of outlier proportion. We first use an initial $\varPhi'=(\varphi_0',\varphi_{-1}',\varphi_{1}')=(N,0,0)$, i.e. $\zeta_0=1$ and $\zeta_{-1}=\zeta_{1}=0$ in $\rho_E(e)$, and train the model by Algorithm \ref{hqrmee}, which means that we expect all samples in the training dataset to achieve minor errors. This will give a resultant model parameter $\omega$, by which we can obtain $\{e_i \}_{i=1}^{N}$ belonging to the continuous interval $(-1,1)$. Then we estimate the outlier proportion by assuming the correctly predicted samples, whose errors belong to $(-0.5,0.5)$, are inliers. On the other hand, the errors belonging to $(-1,-0.5)$ and $(0.5,1)$ correspond to \emph{FN} and \emph{FP} outliers, respectively. Formally, we have
\begin{equation}
\begin{split}
&\varphi_0''=\#\left\{ e_i\in \left( -0.5,0.5 \right) \right\} \; \\ \label{phipara}
&\varphi_{-1}''=\#\left\{ e_i\in \left( -1,-0.5 \right) \right\} \; \\
&\varphi_1''=\#\left\{ e_i\in \left( 0.5,1 \right) \right\} \; \\
\end{split}
\end{equation}
where $\#\left\{ \cdot \right\} $ indicates counting the samples that meet the condition. Obviously we have $\varphi_0''+\varphi_{-1}''+\varphi_1''=N$. With the new $\varPhi''=(\varphi_0'',\varphi_{-1}'',\varphi_{1}'')$, train the model again by Algorithm \ref{hqrmee} and obtain the result of RMEE. 

The above procedure is in fact adaptive. When the training dataset does not contain outliers, it can be supposed that almost all sample are classified well, which means $\varphi_{-1}''$ and $\varphi_1''$ will be of small values. Thus in the following training with $\varPhi''$, we will still expect almost all examples to achieve zero errors. On the other hand, if there are outliers in training dataset, considerable errors will be outside $\left( -0.5,0.5 \right)$. Then, $\varphi_{-1}''$ and $\varphi_1''$ can reflect the outlier proportion to some extent, since higher outlier proportion will generally lead to worse training results, i.e. larger $\varphi_{-1}''$ and $\varphi_1''$.

In addition, note that RMEE with the initial weights $\varPhi'=(\varphi_0',\varphi_{-1}',\varphi_{1}')=(N,0,0)$ is actually equivalent to the C-Loss, the state-of-the-art loss function for robust classification \cite{xu2016robust,ren2018correntropy,singh2014c}, which was proposed from the famous \emph{correntropy} in ITL and aims to maximize the density of $f_E(e)$ at $e=0$. Hence, the proposed RMEE can be regarded as a more generalized form of C-Loss.

\section{Experiments}
\label{sec6}
For performance comparison, above all, it is principal to compare RMEE with QMEE to demonstrate necessity of the proposed restriction. Furthermore, we involve the C-Loss in comparison, which can be viewed as a special case of RMEE, when $\varPhi=(\varphi_0,\varphi_{-1},\varphi_{1})=(N,0,0)$. In addition, the traditional CE and MSE are involved as well. Note that MSE is used with the \emph{sigmoid} transformation. In order to solve the non-convexity problem caused by the use of MSE and \emph{sigmoid} together, the advanced \emph{Adam} \cite{kingma2014adam} is used.

One may probably worry that there are too few algorithms for performance comparison. We would like to argue that:

\noindent \textbf{1}: As reviewed in Section \ref{sec1}, there are a variety of approaches to realize robust classification, such as removing samples, relabeling samples, weighting samples, etc. What most of these methods have in common is that, the desired robustness is realized in the preprocessing stage before the model learns, rather than in learning processes. The proposed RMEE in this paper is a robust objective function for classification, which means RMEE realizes robustness exactly in the learning process. Therefore, it is not necessary to compare RMEE with those methods that achieve robustness outside the learning process. Even RMEE can be used with these methods together.

\noindent \textbf{2}: What should exactly be compared with RMEE are those robust objective functions for classification, among which C-Loss has been
proved to be state-of-the-art. Unlike traditional bounded losses, which are usually truncated by hard threshold \cite{wu2007robust,yang2014robust}, C-Loss is always differentiable, and its kernel size could realize adaptive approximation to various norms in different ranges. For comparisons between C-Loss and existing robust losses, see \cite{xu2016robust,ren2018correntropy,singh2014c}.

The performance indicator through this paper is the classification accuracy that is computed as \emph{(TP+TN)/(TP+TN+FP+FN)}. All the average accuracy and corresponding standard deviations are given by 100 Monte-Carlo independent repetitions. Note that as suggested in \cite{zhu2004class}, to evaluate the robustness between different classifiers, it is preferable to only contaminate training dataset with outliers, while keeping the testing dataset from contamination. This policy has been widely recognized and practiced in the literature of robust machine learning. Therefore, in this paper, the outlier contamination is only aimed at the training datasets, while the testing datasets are unchanged.

\subsection{Logistic Regression}
We first generate linear toy examples which will be contaminated by attribute and label outliers, respectively, and then evaluate the robustness of logistic regression models based on different criteria. Similarly as in \cite{feng2014robust}, we randomly produce 1,000 i.i.d. $x_i\sim \mathcal{N} (0,I_d)$ as training samples, 1,000 i.i.d. $x_i\sim \mathcal{N} (0,I_d)$ as testing samples, and a true solution $\omega^{*} \sim \mathcal{N} (0,I_d)$, where $I_d$ is the unit matrix of dimension $d$. Then, for all the samples, the labels are assigned with $1$ if ${\omega^*}' x_i \geqslant 0$, otherwise with $0$. The dimension is $20$, and all dimensions are relative to the task. The numbers of two classes are supposed to be equal because $\omega^{*}$ always passes through the center of symmetrical Gaussian-distributed samples. As a result, a rather clean dataset is completed.

\subsubsection{Attribute Contamination}
\begin{figure*}[t!]
	\centering
	\includegraphics[width=0.98\textwidth]{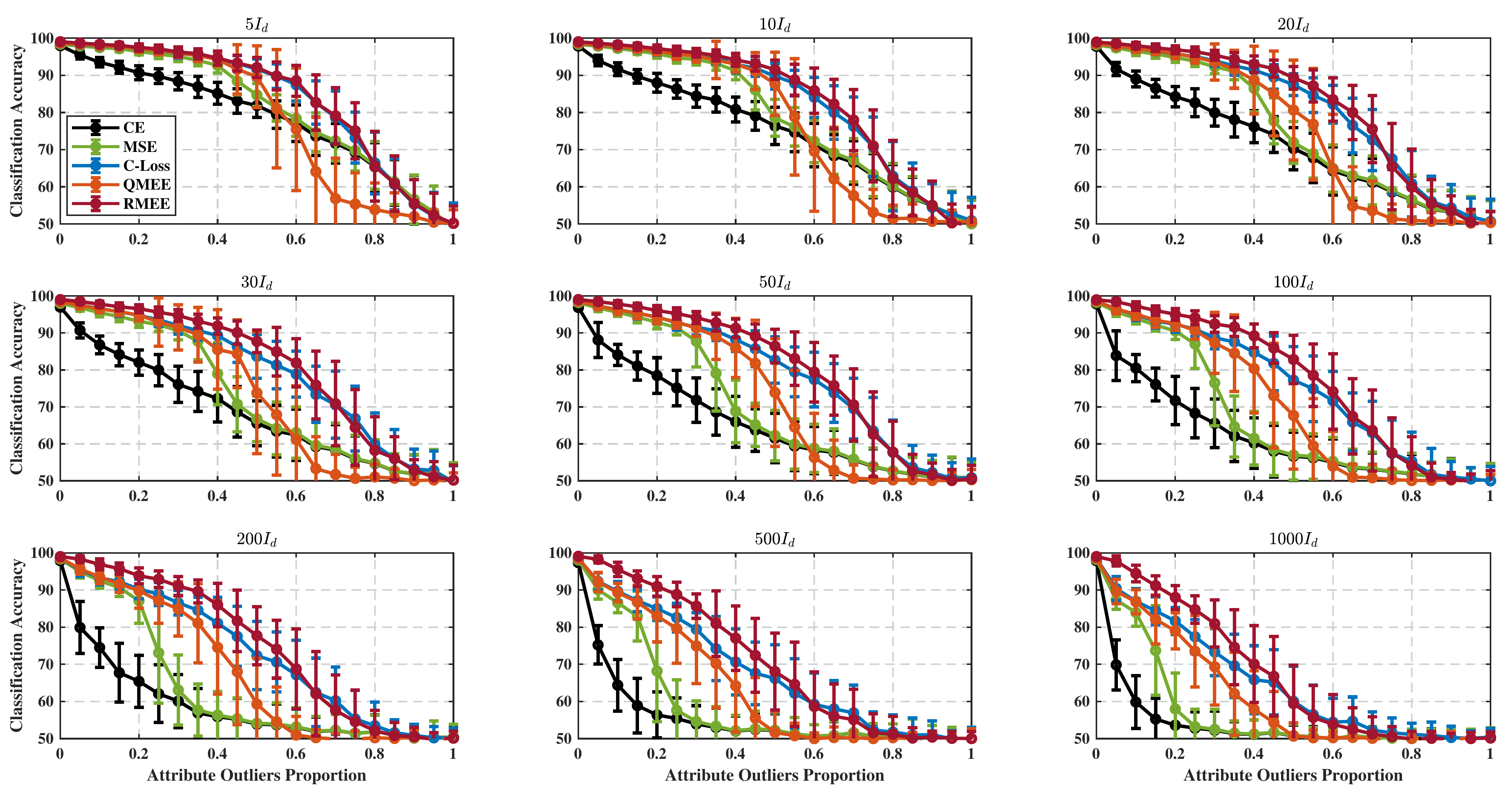}
	\caption{Average classification accuracy of the balanced-class toy dataset contaminated by attribute outliers.}
	\label{fig_attri_out_toy}
\end{figure*}

Generally speaking, attribute contamination has no tendency for different classes since it usually occurs during the measurement process \cite{zhu2004class}. Therefore, the samples of two classes will sustain attribute contamination with equal probability. To contaminate this toy with attribute outliers, we randomly select some samples from the 1,000 training samples, and then replace their attribute values with a zero-mean Gaussian distribution with large covariance to simulate attribute outliers. For the covariance of Gaussian distribution for outliers, we consider several values, which are $5I_d$, $10I_d$, $20I_d$, $30I_d$, $50I_d$, $100I_d$, $200I_d$, $500I_d$, and $1000I_d$, respectively. The number of attribute outliers is denoted by outlier proportion, which is the ratio between the numbers of outliers and total training samples. It is appreciable that when outlier proportion is $1.0$, the accuracy will decrease to chance level because training samples do not carry any valid information. We increase the outlier proportion from $0$ to $1.0$ with a step $0.05$. The results are plotted in Fig. \ref{fig_attri_out_toy}, where one can clearly observe that RMEE achieves the highest accuracy under almost all conditions, which highlights the superiority of MEE for robust classification when restricted by the predetermined codebook.

\subsubsection{Label Contamination}
Compared to attribute contamination, as stated in \cite{frenay2013classification}, samples from certain class could suffer label contamination with more probability. For example, control subjects in medical studies are more prone to be mislabeled \cite{rantalainen2011accounting}. In this paper, we use an asymmetric contamination that, in binary classification, only the samples of one class will be mislabeled whereas those of the other class will not, which requires stronger robustness \cite{zhu2004class,frenay2013classification}. Moreover, in those cases prone to label contamination, the numbers of different classes are often unbalanced. Therefore, besides the above balanced-class toy dataset, we consider unbalanced-class case as well. To generate an unbalanced toy, a similar method as above is used, except that the mean of $x_i$ is shifted to $0.4$ instead, by which the amount of major class vs minor class is $(629.2\pm 86.9)$:$(370.8\pm 86.9)$. 

\begin{figure*}[t!]
	\centering
	\includegraphics[width=0.98\textwidth]{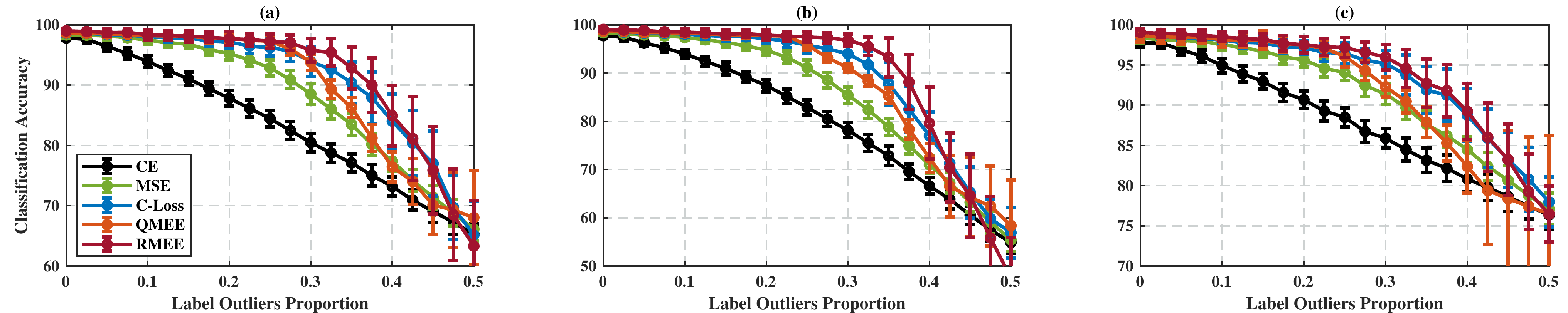}
	\caption{Average classification accuracy of the balanced-class and unbalanced-class toy datasets contaminated by label outliers (a) balanced-class toy, (b) unbalanced-class toy $Maj\rightarrow Min$, (c) unbalanced-class toy $Min\rightarrow Maj$.}
	\label{fig_label_out_toy}
\end{figure*}

For label outlier proportion, we use $Maj\rightarrow Min$ to denote that the labels of major-class samples flip to minor class, and $Min\rightarrow Maj$ for vice versa. First, for the balanced-class toy dataset, we increase the outlier proportion from $0$ to $0.5$ with a step $0.025$. Note that $Maj\rightarrow Min$ and $Min\rightarrow Maj$ are equivalent due to class balance. Next, for the unbalanced-class one, we increase $Maj\rightarrow Min$ or $Min\rightarrow Maj$ from $0$ to $0.5$ with a step $0.025$ bi-directionally while keeping the other as 0. The results are shown in Fig. \ref{fig_label_out_toy}, where one can find that RMEE achieves the highest accuracy in most cases as well.

\subsection{Extreme Learning Machine}
We then select some popular benchmark datasets from the UCI repository \cite{asuncion2007uci} and contaminate them artificially with attribute outliers and label outliers, respectively. The selected datasets are summarized in TABLE \ref{ben_sum}. For each dataset, all the attributes only consist of numerical values. In the context of binary classification, we transform those multi-class datasets into several 2-class datasets. To realize this transformation, we build a new dataset that consists of the samples of one specific class, and assign the antagonistic label to the other samples. Thus, a dataset of $m$ classes is converted into $m$ datasets of binary class, which is known as \emph{one vs all}. This helps analyze whether the classifier could extract effective pattern for each class. We randomly select $2/3$ samples for training, and the other $1/3$ samples act as testing samples.

\begin{table}[h]
	\centering
	\caption{Benchmark datasets summary.}
	\label{ben_sum}
	{
		\begin{tabular}{@{}cccc@{}} 
			\toprule
			\hline
			No. & Dataset & Feature & Class Ratio  \\ \hline
			1& \emph{Statlog (Australian Credit Approval)} &14 & 383 : 307 \\   
			2& \emph{Balance Scale (l. vs all)} &4 & 337 : 288 \\   
			3& \emph{Balance Scale (r. vs all)} &4 & 337 : 288 \\   
			4& \emph{BUPA Liver Disorders}&6 & 200 : 145 \\   
			5& \emph{Connectionist (Sonar, Mines vs. Rocks)} &60 & 111 : 97 \\   
			6& \emph{Iris (set. vs all)} &4 & 100 : 50 \\   
			7& \emph{Iris (vir. vs all)} &4 & 100 : 50 \\   
			8& \emph{Breast Cancer Wisconsin (Original)}&9 & 458 : 241 \\   
			9& \emph{Breast Cancer Wisconsin (Diagnostic)}&30 & 357 : 212 \\   
			10& \emph{Wholesale Customers}&7 & 298 : 142 \\ \hline \bottomrule
	\end{tabular}}
\end{table}

In this subsection, for contaminated benchmark datasets, we use the extreme learning machine (ELM) \cite{huang2006extreme} model for performance evaluation, which is a supervised single-hidden-layer neural network and initializes input weights and hidden layer biases randomly. The robust variant of ELM based on C-Loss was proposed in \cite{ren2018correntropy}. The number of nodes in the hidden layer is set as 50 in this paper.

\subsubsection{Attribute Contamination}
For training samples in each dataset, we first normalize each dimension to zero-mean and unit-variance, so that the diagonal elements of the covariance matrix of the training samples are all 1. Then, similarly, we randomly select some samples and replace their attribute values with a zero-mean Gaussian distribution with large covariance, which are $5I_d$, $20I_d$, $50I_d$, $100I_d$, $300I_d$, and $1000I_d$, respectively. Same as before, the ratio between numbers of outliers and entirety is denoted by the proportion, and the results are listed in TABLE \ref{tab_ben_att}. The highest accuracy in each condition is marked in bold. The `Acoustic' means that the training data is not contaminated in any way.

\begin{table*}[t!]
	\centering
	\caption{Average classification accuracy of benchmark datasets contaminated by attribute outliers.}
	\label{tab_ben_att}
	\resizebox{1.0\textwidth}{!}{
		\begin{tabular}{c c p{1cm}<{\centering}  p{1cm}<{\centering} p{1cm}<{\centering} p{1cm}<{\centering} p{1cm}<{\centering} c c p{1cm}<{\centering}  p{1cm}<{\centering} p{1cm}<{\centering} p{1cm}<{\centering} p{1cm}<{\centering}}
			\toprule
			\hline
			\multicolumn{2}{|c|}{Dataset 1}&\multicolumn{1}{c|}{CE}&\multicolumn{1}{c|}{MSE}&\multicolumn{1}{c|}{C-Loss}&\multicolumn{1}{c|}{QMEE}&\multicolumn{1}{c|}{RMEE}&
			\multicolumn{2}{|c|}{Dataset 2}&\multicolumn{1}{c|}{CE}&\multicolumn{1}{c|}{MSE}&\multicolumn{1}{c|}{C-Loss}&\multicolumn{1}{c|}{QMEE}&\multicolumn{1}{c|}{RMEE}\\ 
			\hline
			\multicolumn{2}{|c|}{Acoustic}&85.3412&85.4808&86.0922&85.1921&\multicolumn{1}{c|}{\textbf{86.8995}}&
			\multicolumn{2}{|c|}{Acoustic}&\textbf{94.6905}&94.6811&94.2130&93.8143&\multicolumn{1}{c|}{94.2067}\\ 
			\hline
			\multicolumn{1}{|c|}{\multirow{2}{*}{$5I_d$}}&\multicolumn{1}{c|}{20\%}&84.6691&84.3992&85.8602&85.1454&\multicolumn{1}{c|}{\textbf{86.6462}}&
			\multicolumn{1}{|c|}{\multirow{2}{*}{$5I_d$}}&\multicolumn{1}{c|}{20\%}&92.3706&94.6656&\textbf{94.9222}&93.4903&\multicolumn{1}{c|}{94.6587}\\ 
			\multicolumn{1}{|c|}{\multirow{2}{*}{}}&\multicolumn{1}{c|}{40\%}&83.9640&83.9368&85.7641&76.5939&\multicolumn{1}{c|}{\textbf{86.0349}}&
			\multicolumn{1}{|c|}{\multirow{2}{*}{}}&\multicolumn{1}{c|}{40\%}&89.3256&91.8122&92.5345&86.3221&\multicolumn{1}{c|}{\textbf{93.6106}}\\ 
			\hline
			\multicolumn{1}{|c|}{\multirow{2}{*}{$20I_d$}}&\multicolumn{1}{c|}{20\%}&84.4759&84.6244&85.9563&81.5240&\multicolumn{1}{c|}{\textbf{86.4890}}&
			\multicolumn{1}{|c|}{\multirow{2}{*}{$20I_d$}}&\multicolumn{1}{c|}{20\%}&91.8823&94.5962&93.9991&90.0144&\multicolumn{1}{c|}{\textbf{94.9760}}\\ 
			\multicolumn{1}{|c|}{\multirow{2}{*}{}}&\multicolumn{1}{c|}{40\%}&83.5499&84.2607&84.5717&77.4716&\multicolumn{1}{c|}{\textbf{85.8427}}&
			\multicolumn{1}{|c|}{\multirow{2}{*}{}}&\multicolumn{1}{c|}{40\%}&87.0992&91.6451&92.4876&77.5385&\multicolumn{1}{c|}{\textbf{92.8413}}\\ 
			\hline
			\multicolumn{1}{|c|}{\multirow{2}{*}{$50I_d$}}&\multicolumn{1}{c|}{20\%}&84.3015&84.7896&\textbf{85.7872}&80.3843&\multicolumn{1}{c|}{85.7118}&
			\multicolumn{1}{|c|}{\multirow{2}{*}{$50I_d$}}&\multicolumn{1}{c|}{20\%}&90.2872&94.0074&93.6400&86.6779&\multicolumn{1}{c|}{\textbf{94.5144}}\\ 
			\multicolumn{1}{|c|}{\multirow{2}{*}{}}&\multicolumn{1}{c|}{40\%}&83.2894&84.8434&84.8872&75.1528&\multicolumn{1}{c|}{\textbf{85.0655}}&  
			\multicolumn{1}{|c|}{\multirow{2}{*}{}}&\multicolumn{1}{c|}{40\%}&83.4484&89.7837&91.1139&74.0913&\multicolumn{1}{c|}{\textbf{92.0048}}\\ 
			\hline
			\multicolumn{1}{|c|}{\multirow{2}{*}{$100I_d$}}&\multicolumn{1}{c|}{20\%}&84.5147&84.9551&\textbf{85.9757}&80.8035&\multicolumn{1}{c|}{85.8472}&
			\multicolumn{1}{|c|}{\multirow{2}{*}{$100I_d$}}&\multicolumn{1}{c|}{20\%}&88.9504&93.5338&93.8302&84.0377&\multicolumn{1}{c|}{\textbf{94.6010}}\\ 
			\multicolumn{1}{|c|}{\multirow{2}{*}{}}&\multicolumn{1}{c|}{40\%}&81.1712&84.3196&84.3645&73.2052&\multicolumn{1}{c|}{\textbf{85.0262}}&
			\multicolumn{1}{|c|}{\multirow{2}{*}{}}&\multicolumn{1}{c|}{40\%}&76.5753&85.2663&88.8492&71.5096&\multicolumn{1}{c|}{\textbf{89.5481}}\\ 
			\hline
			\multicolumn{1}{|c|}{\multirow{2}{*}{$300I_d$}}&\multicolumn{1}{c|}{20\%}&83.2600&84.9286&85.0182&75.5808&\multicolumn{1}{c|}{\textbf{86.0454}}&
			\multicolumn{1}{|c|}{\multirow{2}{*}{$300I_d$}}&\multicolumn{1}{c|}{20\%}&81.7947&91.5097&91.9393&80.3317&\multicolumn{1}{c|}{\textbf{93.5769}}\\ 
			\multicolumn{1}{|c|}{\multirow{2}{*}{}}&\multicolumn{1}{c|}{40\%}&73.2197&82.7930&82.2569&70.8908&\multicolumn{1}{c|}{\textbf{84.0699}}&
			\multicolumn{1}{|c|}{\multirow{2}{*}{}}&\multicolumn{1}{c|}{40\%}&63.4042&71.5518&82.9603&66.6971&\multicolumn{1}{c|}{\textbf{83.3269}}\\ 
			\hline
			\multicolumn{1}{|c|}{\multirow{2}{*}{$1000I_d$}}&\multicolumn{1}{c|}{20\%}&79.3717&84.2168&84.4941&75.3688&\multicolumn{1}{c|}{\textbf{85.3974}}&
			\multicolumn{1}{|c|}{\multirow{2}{*}{$1000I_d$}}&\multicolumn{1}{c|}{20\%}&66.6864&83.0303&87.3118&75.6346&\multicolumn{1}{c|}{\textbf{90.0721}}\\ 
			\multicolumn{1}{|c|}{\multirow{2}{*}{}}&\multicolumn{1}{c|}{40\%}&60.6157&73.1882&77.2149&65.3188&\multicolumn{1}{c|}{\textbf{77.8079}}&
			\multicolumn{1}{|c|}{\multirow{2}{*}{}}&\multicolumn{1}{c|}{40\%}&55.4845&57.0517&67.0551&60.8702&\multicolumn{1}{c|}{\textbf{67.9327}}\\ 
			\hline
			\multicolumn{2}{|c|}{Dataset 3}&\multicolumn{1}{c|}{CE}&\multicolumn{1}{c|}{MSE}&\multicolumn{1}{c|}{C-Loss}&\multicolumn{1}{c|}{QMEE}&\multicolumn{1}{c|}{RMEE}&
			\multicolumn{2}{|c|}{Dataset 4}&\multicolumn{1}{c|}{CE}&\multicolumn{1}{c|}{MSE}&\multicolumn{1}{c|}{C-Loss}&\multicolumn{1}{c|}{QMEE}&\multicolumn{1}{c|}{RMEE}\\ 
			\hline
			\multicolumn{2}{|c|}{Acoustic}&94.9343&94.9031&95.0319&94.3782&\multicolumn{1}{c|}{\textbf{95.1202}}&
			\multicolumn{2}{|c|}{Acoustic}&72.7316&72.2560&72.5066&68.0965&\multicolumn{1}{c|}{\textbf{72.8596}}\\ 
			\hline
			\multicolumn{1}{|c|}{\multirow{2}{*}{$5I_d$}}&\multicolumn{1}{c|}{20\%}&92.1616&94.5170&94.4439&93.3189&\multicolumn{1}{c|}{\textbf{94.9327}}&
			\multicolumn{1}{|c|}{\multirow{2}{*}{$5I_d$}}&\multicolumn{1}{c|}{20\%}&64.7649&67.1150&67.5085&63.5404&\multicolumn{1}{c|}{\textbf{69.8596}}\\ 
			\multicolumn{1}{|c|}{\multirow{2}{*}{}}&\multicolumn{1}{c|}{40\%}&88.7445&91.4719&92.1872&83.9432&\multicolumn{1}{c|}{\textbf{93.1346}}&
			\multicolumn{1}{|c|}{\multirow{2}{*}{}}&\multicolumn{1}{c|}{40\%}&61.5798&62.4558&62.7829&58.5088&\multicolumn{1}{c|}{\textbf{65.1579}}\\ 
			\hline
			\multicolumn{1}{|c|}{\multirow{2}{*}{$20I_d$}}&\multicolumn{1}{c|}{20\%}&92.0175&94.2866&93.7551&93.0048&\multicolumn{1}{c|}{\textbf{94.6971}}&
			\multicolumn{1}{|c|}{\multirow{2}{*}{$20I_d$}}&\multicolumn{1}{c|}{20\%}&64.2705&67.0410&67.2073&63.7105&\multicolumn{1}{c|}{\textbf{67.7368}}\\ 
			\multicolumn{1}{|c|}{\multirow{2}{*}{}}&\multicolumn{1}{c|}{40\%}&86.8134&91.5106&92.3579&81.9183&\multicolumn{1}{c|}{\textbf{92.9808}}&
			\multicolumn{1}{|c|}{\multirow{2}{*}{}}&\multicolumn{1}{c|}{40\%}&59.3699&61.0136&59.9025&56.8158&\multicolumn{1}{c|}{\textbf{61.3070}}\\ 
			\hline
			\multicolumn{1}{|c|}{\multirow{2}{*}{$50I_d$}}&\multicolumn{1}{c|}{20\%}&90.8500&93.3372&93.3186&90.9663&\multicolumn{1}{c|}{\textbf{94.0577}}&
			\multicolumn{1}{|c|}{\multirow{2}{*}{$50I_d$}}&\multicolumn{1}{c|}{20\%}&62.1117&\textbf{65.3978}&65.0258&61.7895&\multicolumn{1}{c|}{65.2281}\\ 
			\multicolumn{1}{|c|}{\multirow{2}{*}{}}&\multicolumn{1}{c|}{40\%}&82.4368&90.0990&91.0431&78.2452&\multicolumn{1}{c|}{\textbf{91.9567}}&
			\multicolumn{1}{|c|}{\multirow{2}{*}{}}&\multicolumn{1}{c|}{40\%}&57.9555&58.6776&58.7636&56.8333&\multicolumn{1}{c|}{\textbf{59.7544}}\\ 
			\hline
			\multicolumn{1}{|c|}{\multirow{2}{*}{$100I_d$}}&\multicolumn{1}{c|}{20\%}&89.1367&93.4503&93.4003&87.6058&\multicolumn{1}{c|}{\textbf{93.8798}}&
			\multicolumn{1}{|c|}{\multirow{2}{*}{$100I_d$}}&\multicolumn{1}{c|}{20\%}&59.6345&62.0637&61.7865&59.2105&\multicolumn{1}{c|}{\textbf{62.1842}}\\ 
			\multicolumn{1}{|c|}{\multirow{2}{*}{}}&\multicolumn{1}{c|}{40\%}&75.1375&85.3168&89.7191&76.7692&\multicolumn{1}{c|}{\textbf{90.1202}}&
			\multicolumn{1}{|c|}{\multirow{2}{*}{}}&\multicolumn{1}{c|}{40\%}&57.7224&58.0879&58.0280&53.6316&\multicolumn{1}{c|}{\textbf{58.5526}}\\ 
			\hline
			\multicolumn{1}{|c|}{\multirow{2}{*}{$300I_d$}}&\multicolumn{1}{c|}{20\%}&81.1232&91.3057&92.1224&83.0962&\multicolumn{1}{c|}{\textbf{93.3846}}&
			\multicolumn{1}{|c|}{\multirow{2}{*}{$300I_d$}}&\multicolumn{1}{c|}{20\%}&57.5348&59.0547&\textbf{59.1582}&58.7632&\multicolumn{1}{c|}{58.8772}\\ 
			\multicolumn{1}{|c|}{\multirow{2}{*}{}}&\multicolumn{1}{c|}{40\%}&63.2575&71.6601&80.7237&70.5961&\multicolumn{1}{c|}{\textbf{81.1827}}&
			\multicolumn{1}{|c|}{\multirow{2}{*}{}}&\multicolumn{1}{c|}{40\%}&57.6294&\textbf{57.7364}&57.0454&53.4649&\multicolumn{1}{c|}{57.7018}\\ 
			\hline
			\multicolumn{1}{|c|}{\multirow{2}{*}{$1000I_d$}}&\multicolumn{1}{c|}{20\%}&65.9159&81.1820&88.5470&80.6346&\multicolumn{1}{c|}{\textbf{89.6490}}&
			\multicolumn{1}{|c|}{\multirow{2}{*}{$1000I_d$}}&\multicolumn{1}{c|}{20\%}&57.6489&\textbf{58.1530}&58.1196&54.6404&\multicolumn{1}{c|}{58.1053}\\ 
			\multicolumn{1}{|c|}{\multirow{2}{*}{}}&\multicolumn{1}{c|}{40\%}&55.1541&57.4967&65.4652&60.8269&\multicolumn{1}{c|}{\textbf{65.7644}}&
			\multicolumn{1}{|c|}{\multirow{2}{*}{}}&\multicolumn{1}{c|}{40\%}&57.7166&\textbf{57.7618}&57.3507&52.5263&\multicolumn{1}{c|}{57.6053}\\ 
			\hline
			\multicolumn{2}{|c|}{Dataset 5}&\multicolumn{1}{c|}{CE}&\multicolumn{1}{c|}{MSE}&\multicolumn{1}{c|}{C-Loss}&\multicolumn{1}{c|}{QMEE}&\multicolumn{1}{c|}{RMEE}&
			\multicolumn{2}{|c|}{Dataset 6}&\multicolumn{1}{c|}{CE}&\multicolumn{1}{c|}{MSE}&\multicolumn{1}{c|}{C-Loss}&\multicolumn{1}{c|}{QMEE}&\multicolumn{1}{c|}{RMEE}\\ 
			\hline
			\multicolumn{2}{|c|}{Acoustic}&77.0282&77.5908&\textbf{77.7576}&77.3712&\multicolumn{1}{c|}{77.5797}&
			\multicolumn{2}{|c|}{Acoustic}&\textbf{99.7893}&99.2746&99.5171&99.0631&\multicolumn{1}{c|}{99.0408}\\ 
			\hline
			\multicolumn{1}{|c|}{\multirow{2}{*}{$5I_d$}}&\multicolumn{1}{c|}{20\%}&73.7599&74.1754&75.8732&70.6087&\multicolumn{1}{c|}{\textbf{75.9565}}&
			\multicolumn{1}{|c|}{\multirow{2}{*}{$5I_d$}}&\multicolumn{1}{c|}{20\%}&\textbf{99.7838}&99.3795&98.7406&97.5918&\multicolumn{1}{c|}{98.5306}\\ 
			\multicolumn{1}{|c|}{\multirow{2}{*}{}}&\multicolumn{1}{c|}{40\%}&71.9547&70.0328&74.0929&69.9565&\multicolumn{1}{c|}{\textbf{74.9130}}&
			\multicolumn{1}{|c|}{\multirow{2}{*}{}}&\multicolumn{1}{c|}{40\%}&\textbf{99.5855}&98.1407&98.5211&87.7347&\multicolumn{1}{c|}{98.6327}\\ 
			\hline
			\multicolumn{1}{|c|}{\multirow{2}{*}{$20I_d$}}&\multicolumn{1}{c|}{20\%}&72.8445&72.7313&74.1694&70.1594&\multicolumn{1}{c|}{\textbf{75.8841}}&
			\multicolumn{1}{|c|}{\multirow{2}{*}{$20I_d$}}&\multicolumn{1}{c|}{20\%}&\textbf{99.6234}&99.1297&99.0156&96.7143&\multicolumn{1}{c|}{97.6327}\\ 
			\multicolumn{1}{|c|}{\multirow{2}{*}{}}&\multicolumn{1}{c|}{40\%}&70.9818&70.0738&\textbf{74.6192}&69.1739&\multicolumn{1}{c|}{74.3188}&
			\multicolumn{1}{|c|}{\multirow{2}{*}{}}&\multicolumn{1}{c|}{40\%}&\textbf{99.4411}&98.6793&97.0664&86.3061&\multicolumn{1}{c|}{96.1224}\\ 
			\hline
			\multicolumn{1}{|c|}{\multirow{2}{*}{$50I_d$}}&\multicolumn{1}{c|}{20\%}&72.2072&72.2641&74.2687&70.9275&\multicolumn{1}{c|}{\textbf{75.7681}}&
			\multicolumn{1}{|c|}{\multirow{2}{*}{$50I_d$}}&\multicolumn{1}{c|}{20\%}&\textbf{99.7710}&98.8713&98.2464&95.6939&\multicolumn{1}{c|}{97.9592}\\ 
			\multicolumn{1}{|c|}{\multirow{2}{*}{}}&\multicolumn{1}{c|}{40\%}&70.9769&70.9091&74.1996&68.6087&\multicolumn{1}{c|}{\textbf{74.8116}}&
			\multicolumn{1}{|c|}{\multirow{2}{*}{}}&\multicolumn{1}{c|}{40\%}&97.1245&\textbf{98.5852}&97.7731&85.3469&\multicolumn{1}{c|}{95.6939}\\ 
			\hline
			\multicolumn{1}{|c|}{\multirow{2}{*}{$100I_d$}}&\multicolumn{1}{c|}{20\%}&72.7967&72.5096&74.2348&71.8441&\multicolumn{1}{c|}{\textbf{75.4783}}&
			\multicolumn{1}{|c|}{\multirow{2}{*}{$100I_d$}}&\multicolumn{1}{c|}{20\%}&\textbf{99.7606}&98.1226&97.0511&93.9388&\multicolumn{1}{c|}{97.2857}\\ 
			\multicolumn{1}{|c|}{\multirow{2}{*}{}}&\multicolumn{1}{c|}{40\%}&69.3462&70.1595&73.8919&68.0870&\multicolumn{1}{c|}{\textbf{74.3333}}&
			\multicolumn{1}{|c|}{\multirow{2}{*}{}}&\multicolumn{1}{c|}{40\%}&91.8587&\textbf{98.1526}&97.4333&84.9388&\multicolumn{1}{c|}{96.0612}\\ 
			\hline
			\multicolumn{1}{|c|}{\multirow{2}{*}{$300I_d$}}&\multicolumn{1}{c|}{20\%}&70.4031&70.7418&72.6084&70.6232&\multicolumn{1}{c|}{\textbf{73.6522}}&
			\multicolumn{1}{|c|}{\multirow{2}{*}{$300I_d$}}&\multicolumn{1}{c|}{20\%}&97.7551&\textbf{98.8015}&97.7083&89.3265&\multicolumn{1}{c|}{97.6122}\\ 
			\multicolumn{1}{|c|}{\multirow{2}{*}{}}&\multicolumn{1}{c|}{40\%}&68.0087&70.4738&\textbf{72.5308}&66.6973&\multicolumn{1}{c|}{71.6377}&
			\multicolumn{1}{|c|}{\multirow{2}{*}{}}&\multicolumn{1}{c|}{40\%}&73.3285&90.4130&93.9672&83.9388&\multicolumn{1}{c|}{\textbf{95.2449}}\\ 
			\hline
			\multicolumn{1}{|c|}{\multirow{2}{*}{$1000I_d$}}&\multicolumn{1}{c|}{20\%}&70.8000&70.4146&\textbf{72.9791}&67.7681&\multicolumn{1}{c|}{72.1159}&
			\multicolumn{1}{|c|}{\multirow{2}{*}{$1000I_d$}}&\multicolumn{1}{c|}{20\%}&83.2386&\textbf{98.1099}&96.6321&89.4286&\multicolumn{1}{c|}{94.6735}\\ 
			\multicolumn{1}{|c|}{\multirow{2}{*}{}}&\multicolumn{1}{c|}{40\%}&65.3351&69.6401&69.3089&65.1159&\multicolumn{1}{c|}{\textbf{70.3913}}&
			\multicolumn{1}{|c|}{\multirow{2}{*}{}}&\multicolumn{1}{c|}{40\%}&67.9696&75.8694&78.1879&\textbf{79.5714}&\multicolumn{1}{c|}{79.4082}\\ 
			\hline
			\multicolumn{2}{|c|}{Dataset 7}&\multicolumn{1}{c|}{CE}&\multicolumn{1}{c|}{MSE}&\multicolumn{1}{c|}{C-Loss}&\multicolumn{1}{c|}{QMEE}&\multicolumn{1}{c|}{RMEE}&
			\multicolumn{2}{|c|}{Dataset 8}&\multicolumn{1}{c|}{CE}&\multicolumn{1}{c|}{MSE}&\multicolumn{1}{c|}{C-Loss}&\multicolumn{1}{c|}{QMEE}&\multicolumn{1}{c|}{RMEE}\\ 
			\hline
			\multicolumn{2}{|c|}{Acoustic}&\textbf{96.2420}&94.3650&94.7064&95.0000&\multicolumn{1}{c|}{95.0408}&
			\multicolumn{2}{|c|}{Acoustic}&\textbf{96.3070}&95.1221&95.5036&95.5388&\multicolumn{1}{c|}{95.7500}\\ 
			\hline
			\multicolumn{1}{|c|}{\multirow{2}{*}{$5I_d$}}&\multicolumn{1}{c|}{20\%}&93.9750&94.6976&94.3587&94.3854&\multicolumn{1}{c|}{\textbf{95.0776}}&
			\multicolumn{1}{|c|}{\multirow{2}{*}{$5I_d$}}&\multicolumn{1}{c|}{20\%}&\textbf{95.8097}&95.5167&94.4502&94.7414&\multicolumn{1}{c|}{95.4698}\\ 
			\multicolumn{1}{|c|}{\multirow{2}{*}{}}&\multicolumn{1}{c|}{40\%}&89.9350&90.5405&91.9693&85.1633&\multicolumn{1}{c|}{\textbf{92.5510}}&
			\multicolumn{1}{|c|}{\multirow{2}{*}{}}&\multicolumn{1}{c|}{40\%}&95.3661&\textbf{95.6059}&94.7405&94.9009&\multicolumn{1}{c|}{95.5302}\\ 
			\hline
			\multicolumn{1}{|c|}{\multirow{2}{*}{$20I_d$}}&\multicolumn{1}{c|}{20\%}&92.1655&92.6733&92.6554&90.9796&\multicolumn{1}{c|}{\textbf{93.1837}}&
			\multicolumn{1}{|c|}{\multirow{2}{*}{$20I_d$}}&\multicolumn{1}{c|}{20\%}&\textbf{95.9798}&95.8572&94.4977&93.6881&\multicolumn{1}{c|}{95.8534}\\ 
			\multicolumn{1}{|c|}{\multirow{2}{*}{}}&\multicolumn{1}{c|}{40\%}&88.8540&90.6625&90.4983&84.0816&\multicolumn{1}{c|}{\textbf{91.4898}}&
			\multicolumn{1}{|c|}{\multirow{2}{*}{}}&\multicolumn{1}{c|}{40\%}&95.1232&\textbf{95.6049}&94.0459&92.7198&\multicolumn{1}{c|}{94.7716}\\ 
			\hline
			\multicolumn{1}{|c|}{\multirow{2}{*}{$50I_d$}}&\multicolumn{1}{c|}{20\%}&90.4424&91.5963&91.1111&90.6531&\multicolumn{1}{c|}{\textbf{91.9592}}&
			\multicolumn{1}{|c|}{\multirow{2}{*}{$50I_d$}}&\multicolumn{1}{c|}{20\%}&95.1250&94.4612&94.8319&91.4655&\multicolumn{1}{c|}{\textbf{95.1595}}\\ 
			\multicolumn{1}{|c|}{\multirow{2}{*}{}}&\multicolumn{1}{c|}{40\%}&83.6684&85.1672&88.0256&82.5306&\multicolumn{1}{c|}{\textbf{88.2857}}&
			\multicolumn{1}{|c|}{\multirow{2}{*}{}}&\multicolumn{1}{c|}{40\%}&93.2328&\textbf{93.7414}&92.8060&87.6810&\multicolumn{1}{c|}{92.3448}\\ 
			\hline
			\multicolumn{1}{|c|}{\multirow{2}{*}{$100I_d$}}&\multicolumn{1}{c|}{20\%}&89.7892&90.7375&91.5391&88.5714&\multicolumn{1}{c|}{\textbf{91.9388}}&
			\multicolumn{1}{|c|}{\multirow{2}{*}{$100I_d$}}&\multicolumn{1}{c|}{20\%}&94.7328&95.2586&94.0905&88.2026&\multicolumn{1}{c|}{\textbf{95.3362}}\\ 
			\multicolumn{1}{|c|}{\multirow{2}{*}{}}&\multicolumn{1}{c|}{40\%}&77.2352&83.2288&85.2843&80.6351&\multicolumn{1}{c|}{\textbf{85.6939}}&
			\multicolumn{1}{|c|}{\multirow{2}{*}{}}&\multicolumn{1}{c|}{40\%}&90.1379&91.9310&92.2629&81.3966&\multicolumn{1}{c|}{\textbf{92.4543}}\\ 
			\hline
			\multicolumn{1}{|c|}{\multirow{2}{*}{$300I_d$}}&\multicolumn{1}{c|}{20\%}&83.2700&91.2687&90.4733&85.7143&\multicolumn{1}{c|}{\textbf{91.4082}}&
			\multicolumn{1}{|c|}{\multirow{2}{*}{$300I_d$}}&\multicolumn{1}{c|}{20\%}&93.5431&91.9612&93.2586&85.6987&\multicolumn{1}{c|}{\textbf{94.7371}}\\ 
			\multicolumn{1}{|c|}{\multirow{2}{*}{}}&\multicolumn{1}{c|}{40\%}&70.0754&73.2084&80.4062&76.3673&\multicolumn{1}{c|}{\textbf{80.5306}}&
			\multicolumn{1}{|c|}{\multirow{2}{*}{}}&\multicolumn{1}{c|}{40\%}&78.8405&90.1078&91.4957&76.9483&\multicolumn{1}{c|}{\textbf{92.1078}}\\ 
			\hline
			\multicolumn{1}{|c|}{\multirow{2}{*}{$1000I_d$}}&\multicolumn{1}{c|}{20\%}&72.4003&86.0412&86.1430&80.5714&\multicolumn{1}{c|}{\textbf{88.6735}}&
			\multicolumn{1}{|c|}{\multirow{2}{*}{$1000I_d$}}&\multicolumn{1}{c|}{20\%}&84.4741&90.0776&91.6724&80.9914&\multicolumn{1}{c|}{\textbf{93.2543}}\\ 
			\multicolumn{1}{|c|}{\multirow{2}{*}{}}&\multicolumn{1}{c|}{40\%}&67.6567&70.2632&\textbf{73.4066}&70.1020&\multicolumn{1}{c|}{71.2041}&
			\multicolumn{1}{|c|}{\multirow{2}{*}{}}&\multicolumn{1}{c|}{40\%}&65.9052&81.3448&\textbf{83.9138}&73.3750&\multicolumn{1}{c|}{83.3922}\\ 
			\hline
			\multicolumn{2}{|c|}{Dataset 9}&\multicolumn{1}{c|}{CE}&\multicolumn{1}{c|}{MSE}&\multicolumn{1}{c|}{C-Loss}&\multicolumn{1}{c|}{QMEE}&\multicolumn{1}{c|}{RMEE}&
			\multicolumn{2}{|c|}{Dataset 10}&\multicolumn{1}{c|}{CE}&\multicolumn{1}{c|}{MSE}&\multicolumn{1}{c|}{C-Loss}&\multicolumn{1}{c|}{QMEE}&\multicolumn{1}{c|}{RMEE}\\ 
			\hline
			\multicolumn{2}{|c|}{Acoustic}&96.3129&96.0894&95.8960&95.5556&\multicolumn{1}{c|}{\textbf{96.5185}}&
			\multicolumn{2}{|c|}{Acoustic}&\textbf{90.6831}&90.3615&90.2134&89.9932&\multicolumn{1}{c|}{89.9178}\\ 
			\hline
			\multicolumn{1}{|c|}{\multirow{2}{*}{$5I_d$}}&\multicolumn{1}{c|}{20\%}&\textbf{95.7139}&95.2151&94.7992&94.9101&\multicolumn{1}{c|}{95.6878}&
			\multicolumn{1}{|c|}{\multirow{2}{*}{$5I_d$}}&\multicolumn{1}{c|}{20\%}&\textbf{88.1849}&86.8219&87.3904&85.6233&\multicolumn{1}{c|}{87.0616}\\ 
			\multicolumn{1}{|c|}{\multirow{2}{*}{}}&\multicolumn{1}{c|}{40\%}&94.4134&94.0203&93.2595&92.1005&\multicolumn{1}{c|}{\textbf{94.6402}}&
			\multicolumn{1}{|c|}{\multirow{2}{*}{}}&\multicolumn{1}{c|}{40\%}&\textbf{85.4521}&84.3767&84.7192&80.5068&\multicolumn{1}{c|}{84.8836}\\ 
			\hline
			\multicolumn{1}{|c|}{\multirow{2}{*}{$20I_d$}}&\multicolumn{1}{c|}{20\%}&\textbf{95.4883}&95.4513&95.2227&92.0582&\multicolumn{1}{c|}{95.2804}&
			\multicolumn{1}{|c|}{\multirow{2}{*}{$20I_d$}}&\multicolumn{1}{c|}{20\%}&86.9247&86.4315&87.3014&78.4726&\multicolumn{1}{c|}{\textbf{87.4726}}\\ 
			\multicolumn{1}{|c|}{\multirow{2}{*}{}}&\multicolumn{1}{c|}{40\%}&94.3818&94.6093&93.7900&90.1323&\multicolumn{1}{c|}{\textbf{94.7302}}&
			\multicolumn{1}{|c|}{\multirow{2}{*}{}}&\multicolumn{1}{c|}{40\%}&81.1370&\textbf{84.5000}&82.0137&78.2699&\multicolumn{1}{c|}{82.7260}\\ 
			\hline
			\multicolumn{1}{|c|}{\multirow{2}{*}{$50I_d$}}&\multicolumn{1}{c|}{20\%}&95.7556&\textbf{95.8707}&95.8267&91.9418&\multicolumn{1}{c|}{95.4974}&
			\multicolumn{1}{|c|}{\multirow{2}{*}{$50I_d$}}&\multicolumn{1}{c|}{20\%}&84.9041&86.0616&86.0616&77.9041&\multicolumn{1}{c|}{\textbf{86.5411}}\\ 
			\multicolumn{1}{|c|}{\multirow{2}{*}{}}&\multicolumn{1}{c|}{40\%}&93.8356&94.7828&94.6305&90.4021&\multicolumn{1}{c|}{\textbf{94.9577}}&
			\multicolumn{1}{|c|}{\multirow{2}{*}{}}&\multicolumn{1}{c|}{40\%}&76.0068&\textbf{83.4589}&82.5000&73.1575&\multicolumn{1}{c|}{80.0685}\\ 
			\hline
			\multicolumn{1}{|c|}{\multirow{2}{*}{$100I_d$}}&\multicolumn{1}{c|}{20\%}&94.5714&95.2646&95.0688&88.3175&\multicolumn{1}{c|}{\textbf{95.4074}}&
			\multicolumn{1}{|c|}{\multirow{2}{*}{$100I_d$}}&\multicolumn{1}{c|}{20\%}&81.6507&85.3219&\textbf{85.7123}&77.8836&\multicolumn{1}{c|}{85.3767}\\ 
			\multicolumn{1}{|c|}{\multirow{2}{*}{}}&\multicolumn{1}{c|}{40\%}&92.5873&93.3069&94.3175&85.9206&\multicolumn{1}{c|}{\textbf{94.4815}}&
			\multicolumn{1}{|c|}{\multirow{2}{*}{}}&\multicolumn{1}{c|}{40\%}&71.8699&79.4315&80.7123&70.5205&\multicolumn{1}{c|}{\textbf{80.9041}}\\ 
			\hline
			\multicolumn{1}{|c|}{\multirow{2}{*}{$300I_d$}}&\multicolumn{1}{c|}{20\%}&93.1693&94.1005&94.2751&85.4021&\multicolumn{1}{c|}{\textbf{94.6614}}&
			\multicolumn{1}{|c|}{\multirow{2}{*}{$300I_d$}}&\multicolumn{1}{c|}{20\%}&75.5411&81.3288&\textbf{83.5822}&77.4521&\multicolumn{1}{c|}{83.2055}\\ 
			\multicolumn{1}{|c|}{\multirow{2}{*}{}}&\multicolumn{1}{c|}{40\%}&89.3545&92.9153&92.3386&75.6984&\multicolumn{1}{c|}{\textbf{93.8730}}&
			\multicolumn{1}{|c|}{\multirow{2}{*}{}}&\multicolumn{1}{c|}{40\%}&68.6712&72.6233&72.9315&65.4178&\multicolumn{1}{c|}{\textbf{73.0342}}\\ 
			\hline
			\multicolumn{1}{|c|}{\multirow{2}{*}{$1000I_d$}}&\multicolumn{1}{c|}{20\%}&89.8042&92.4021&93.8571&80.1270&\multicolumn{1}{c|}{\textbf{94.7513}}&
			\multicolumn{1}{|c|}{\multirow{2}{*}{$1000I_d$}}&\multicolumn{1}{c|}{20\%}&69.2671&74.3151&77.1438&71.4521&\multicolumn{1}{c|}{\textbf{78.0479}}\\ 
			\multicolumn{1}{|c|}{\multirow{2}{*}{}}&\multicolumn{1}{c|}{40\%}&78.0370&89.8148&90.3810&73.7249&\multicolumn{1}{c|}{\textbf{91.0529}}&
			\multicolumn{1}{|c|}{\multirow{2}{*}{}}&\multicolumn{1}{c|}{40\%}&67.7671&68.0822&68.2123&61.1575&\multicolumn{1}{c|}{\textbf{68.4658}}\\ 
			\hline
			\bottomrule
	\end{tabular}}
\end{table*}

\subsubsection{Label Contamination}
We similarly increase $Maj\rightarrow Min$ or $Min\rightarrow Maj$ bi-directionally while keeping the other as 0. The results are listed in TABLE \ref{tab_ben_lab}, and likewise bold font means the highest accuracy in each case.

\begin{table*}[t]
	\centering
	\caption{Average classification accuracy of benchmark datasets contaminated by label outliers.}
	\label{tab_ben_lab}
	\resizebox{1.0\textwidth}{!}{
		\begin{tabular}{c c p{1cm}<{\centering}  p{1cm}<{\centering} p{1cm}<{\centering} p{1cm}<{\centering} p{1cm}<{\centering} c c p{1cm}<{\centering}  p{1cm}<{\centering} p{1cm}<{\centering} p{1cm}<{\centering} p{1cm}<{\centering}}
			\toprule
			\hline
			\multicolumn{2}{|c|}{Dataset 1}&\multicolumn{1}{c|}{CE}&\multicolumn{1}{c|}{MSE}&\multicolumn{1}{c|}{C-Loss}&\multicolumn{1}{c|}{QMEE}&\multicolumn{1}{c|}{RMEE}&
			\multicolumn{2}{|c|}{Dataset 2}&\multicolumn{1}{c|}{CE}&\multicolumn{1}{c|}{MSE}&\multicolumn{1}{c|}{C-Loss}&\multicolumn{1}{c|}{QMEE}&\multicolumn{1}{c|}{RMEE}\\ 
			\hline
			\multicolumn{1}{|c|}{\multirowcell{3}{$Maj$\\$\downarrow$\\$Min$}}&\multicolumn{1}{c|}{10\%}&84.3474&84.4507&83.8348&83.6725&\multicolumn{1}{c|}{\textbf{86.0000}}&
			\multicolumn{1}{|c|}{\multirowcell{3}{$Maj$\\$\downarrow$\\$Min$}}&\multicolumn{1}{c|}{10\%}&91.7168&94.4214&\textbf{94.7259}&90.1917&\multicolumn{1}{c|}{94.0433}\\ 
			\multicolumn{1}{|c|}{\multirow{3}{*}{}}&\multicolumn{1}{c|}{20\%}&82.1436&81.6331&80.8427&81.8079&\multicolumn{1}{c|}{\textbf{84.9083}}&
			\multicolumn{1}{|c|}{\multirow{3}{*}{}}&\multicolumn{1}{c|}{20\%}&87.6409&91.9176&94.0952&89.2692&\multicolumn{1}{c|}{\textbf{94.3221}}\\ 
			\multicolumn{1}{|c|}{\multirow{3}{*}{}}&\multicolumn{1}{c|}{30\%}&77.9699&77.6036&77.2135&75.7686&\multicolumn{1}{c|}{\textbf{80.9869}}&
			\multicolumn{1}{|c|}{\multirow{3}{*}{}}&\multicolumn{1}{c|}{30\%}&82.5877&83.6988&89.8441&87.0048&\multicolumn{1}{c|}{\textbf{93.3942}}\\
			\hline
			\multicolumn{1}{|c|}{\multirowcell{3}{$Min$\\$\downarrow$\\$Maj$}}&\multicolumn{1}{c|}{10\%}&84.5616&84.1731&83.5567&83.5459&\multicolumn{1}{c|}{\textbf{85.4629}}&
			\multicolumn{1}{|c|}{\multirowcell{3}{$Min$\\$\downarrow$\\$Maj$}}&\multicolumn{1}{c|}{10\%}&93.2490&\textbf{95.1590}&94.3271&89.3413&\multicolumn{1}{c|}{94.4712}\\ 
			\multicolumn{1}{|c|}{\multirow{3}{*}{}}&\multicolumn{1}{c|}{20\%}&82.1527&81.0994&81.5897&78.6332&\multicolumn{1}{c|}{\textbf{83.8646}}&
			\multicolumn{1}{|c|}{\multirow{3}{*}{}}&\multicolumn{1}{c|}{20\%}&88.7199&93.1874&93.8398&88.5000&\multicolumn{1}{c|}{\textbf{93.9856}}\\ 
			\multicolumn{1}{|c|}{\multirow{3}{*}{}}&\multicolumn{1}{c|}{30\%}&\textbf{78.3863}&76.3002&75.8697&76.4541&\multicolumn{1}{c|}{77.3144}&
			\multicolumn{1}{|c|}{\multirow{3}{*}{}}&\multicolumn{1}{c|}{30\%}&83.8488&84.0183&86.3634&86.1779&\multicolumn{1}{c|}{\textbf{87.7260}}\\
			\hline
			\multicolumn{2}{|c|}{Dataset 3}&\multicolumn{1}{c|}{CE}&\multicolumn{1}{c|}{MSE}&\multicolumn{1}{c|}{C-Loss}&\multicolumn{1}{c|}{QMEE}&\multicolumn{1}{c|}{RMEE}&
			\multicolumn{2}{|c|}{Dataset 4}&\multicolumn{1}{c|}{CE}&\multicolumn{1}{c|}{MSE}&\multicolumn{1}{c|}{C-Loss}&\multicolumn{1}{c|}{QMEE}&\multicolumn{1}{c|}{RMEE}\\ 
			\hline
			\multicolumn{1}{|c|}{\multirowcell{3}{$Maj$\\$\downarrow$\\$Min$}}&\multicolumn{1}{c|}{10\%}&91.4893&\textbf{94.7427}&94.1611&93.8558&\multicolumn{1}{c|}{94.1587}&
			\multicolumn{1}{|c|}{\multirowcell{3}{$Maj$\\$\downarrow$\\$Min$}}&\multicolumn{1}{c|}{10\%}&66.8953&68.5372&71.5927&68.2807&\multicolumn{1}{c|}{\textbf{72.0439}}\\ 
			\multicolumn{1}{|c|}{\multirow{3}{*}{}}&\multicolumn{1}{c|}{20\%}&88.1498&92.3192&93.6099&90.5721&\multicolumn{1}{c|}{\textbf{93.8077}}&
			\multicolumn{1}{|c|}{\multirow{3}{*}{}}&\multicolumn{1}{c|}{20\%}&63.2676&64.0844&66.6528&63.8684&\multicolumn{1}{c|}{\textbf{67.2281}}\\ 
			\multicolumn{1}{|c|}{\multirow{3}{*}{}}&\multicolumn{1}{c|}{30\%}&81.3765&84.2534&89.0305&89.1731&\multicolumn{1}{c|}{\textbf{90.8125}}&
			\multicolumn{1}{|c|}{\multirow{3}{*}{}}&\multicolumn{1}{c|}{30\%}&57.4080&58.2489&58.0663&\textbf{60.4649}&\multicolumn{1}{c|}{58.2281}\\
			\hline
			\multicolumn{1}{|c|}{\multirowcell{3}{$Min$\\$\downarrow$\\$Maj$}}&\multicolumn{1}{c|}{10\%}&93.4255&94.1099&93.4726&92.2596&\multicolumn{1}{c|}{\textbf{94.5000}}&
			\multicolumn{1}{|c|}{\multirowcell{3}{$Min$\\$\downarrow$\\$Maj$}}&\multicolumn{1}{c|}{10\%}&69.1810&\textbf{70.2557}&70.0053&66.9211&\multicolumn{1}{c|}{69.9298}\\ 
			\multicolumn{1}{|c|}{\multirow{3}{*}{}}&\multicolumn{1}{c|}{20\%}&88.7899&92.5616&92.4826&90.9606&\multicolumn{1}{c|}{\textbf{93.7933}}&
			\multicolumn{1}{|c|}{\multirow{3}{*}{}}&\multicolumn{1}{c|}{20\%}&68.1855&\textbf{69.0621}&68.3790&65.4474&\multicolumn{1}{c|}{66.0088}\\ 
			\multicolumn{1}{|c|}{\multirow{3}{*}{}}&\multicolumn{1}{c|}{30\%}&83.6452&84.8116&87.5163&\textbf{88.2404}&\multicolumn{1}{c|}{88.0288}&
			\multicolumn{1}{|c|}{\multirow{3}{*}{}}&\multicolumn{1}{c|}{30\%}&65.0746&\textbf{65.3717}&63.9841&63.8684&\multicolumn{1}{c|}{61.7456}\\
			\hline
			\multicolumn{2}{|c|}{Dataset 5}&\multicolumn{1}{c|}{CE}&\multicolumn{1}{c|}{MSE}&\multicolumn{1}{c|}{C-Loss}&\multicolumn{1}{c|}{QMEE}&\multicolumn{1}{c|}{RMEE}&
			\multicolumn{2}{|c|}{Dataset 6}&\multicolumn{1}{c|}{CE}&\multicolumn{1}{c|}{MSE}&\multicolumn{1}{c|}{C-Loss}&\multicolumn{1}{c|}{QMEE}&\multicolumn{1}{c|}{RMEE}\\ 
			\hline
			\multicolumn{1}{|c|}{\multirowcell{3}{$Maj$\\$\downarrow$\\$Min$}}&\multicolumn{1}{c|}{10\%}&71.4133&71.4726&75.2309&72.0290&\multicolumn{1}{c|}{\textbf{76.1159}}&
			\multicolumn{1}{|c|}{\multirowcell{3}{$Maj$\\$\downarrow$\\$Min$}}&\multicolumn{1}{c|}{10\%}&\textbf{99.0180}&98.1857&96.1050&95.6531&\multicolumn{1}{c|}{97.2857}\\ 
			\multicolumn{1}{|c|}{\multirow{3}{*}{}}&\multicolumn{1}{c|}{20\%}&67.5098&67.4807&72.2781&69.1449&\multicolumn{1}{c|}{\textbf{74.6377}}&
			\multicolumn{1}{|c|}{\multirow{3}{*}{}}&\multicolumn{1}{c|}{20\%}&\textbf{97.6726}&96.3080&96.4170&93.1429&\multicolumn{1}{c|}{95.5306}\\ 
			\multicolumn{1}{|c|}{\multirow{3}{*}{}}&\multicolumn{1}{c|}{30\%}&65.3542&65.0629&65.6654&65.5507&\multicolumn{1}{c|}{\textbf{66.9420}}&
			\multicolumn{1}{|c|}{\multirow{3}{*}{}}&\multicolumn{1}{c|}{30\%}&92.5694&93.3054&92.1344&93.0165&\multicolumn{1}{c|}{\textbf{94.0408}}\\
			\hline
			\multicolumn{1}{|c|}{\multirowcell{3}{$Min$\\$\downarrow$\\$Maj$}}&\multicolumn{1}{c|}{10\%}&70.9895&70.9457&75.2593&71.9275&\multicolumn{1}{c|}{\textbf{76.0145}}&
			\multicolumn{1}{|c|}{\multirowcell{3}{$Min$\\$\downarrow$\\$Maj$}}&\multicolumn{1}{c|}{10\%}&\textbf{99.5201}&98.4163&96.8473&93.8980&\multicolumn{1}{c|}{96.3673}\\ 
			\multicolumn{1}{|c|}{\multirow{3}{*}{}}&\multicolumn{1}{c|}{20\%}&69.7249&69.9677&71.9783&69.7101&\multicolumn{1}{c|}{\textbf{72.4058}}&
			\multicolumn{1}{|c|}{\multirow{3}{*}{}}&\multicolumn{1}{c|}{20\%}&\textbf{98.3173}&95.8684&94.1687&92.9388&\multicolumn{1}{c|}{96.2449}\\ 
			\multicolumn{1}{|c|}{\multirow{3}{*}{}}&\multicolumn{1}{c|}{30\%}&67.5924&66.5936&67.6378&\textbf{68.2899}&\multicolumn{1}{c|}{67.4493}&
			\multicolumn{1}{|c|}{\multirow{3}{*}{}}&\multicolumn{1}{c|}{30\%}&\textbf{96.2771}&94.7256&90.8305&91.4694&\multicolumn{1}{c|}{93.0204}\\
			\hline
			\multicolumn{2}{|c|}{Dataset 7}&\multicolumn{1}{c|}{CE}&\multicolumn{1}{c|}{MSE}&\multicolumn{1}{c|}{C-Loss}&\multicolumn{1}{c|}{QMEE}&\multicolumn{1}{c|}{RMEE}&
			\multicolumn{2}{|c|}{Dataset 8}&\multicolumn{1}{c|}{CE}&\multicolumn{1}{c|}{MSE}&\multicolumn{1}{c|}{C-Loss}&\multicolumn{1}{c|}{QMEE}&\multicolumn{1}{c|}{RMEE}\\ 
			\hline
			\multicolumn{1}{|c|}{\multirowcell{3}{$Maj$\\$\downarrow$\\$Min$}}&\multicolumn{1}{c|}{10\%}&93.5932&94.6465&\textbf{95.4148}&92.3061&\multicolumn{1}{c|}{94.5510}&
			\multicolumn{1}{|c|}{\multirowcell{3}{$Maj$\\$\downarrow$\\$Min$}}&\multicolumn{1}{c|}{10\%}&\textbf{96.2773}&94.9510&95.0918&93.8621&\multicolumn{1}{c|}{96.1767}\\ 
			\multicolumn{1}{|c|}{\multirow{3}{*}{}}&\multicolumn{1}{c|}{20\%}&90.6252&93.9163&95.0767&90.5714&\multicolumn{1}{c|}{\textbf{95.4082}}&
			\multicolumn{1}{|c|}{\multirow{3}{*}{}}&\multicolumn{1}{c|}{20\%}&\textbf{95.4819}&94.1410&94.8297&92.7759&\multicolumn{1}{c|}{94.7931}\\ 
			\multicolumn{1}{|c|}{\multirow{3}{*}{}}&\multicolumn{1}{c|}{30\%}&85.1919&87.1789&92.6509&89.8980&\multicolumn{1}{c|}{\textbf{93.9796}}&
			\multicolumn{1}{|c|}{\multirow{3}{*}{}}&\multicolumn{1}{c|}{30\%}&94.0884&93.3471&94.2440&91.9440&\multicolumn{1}{c|}{\textbf{94.5129}}\\
			\hline
			\multicolumn{1}{|c|}{\multirowcell{3}{$Min$\\$\downarrow$\\$Maj$}}&\multicolumn{1}{c|}{10\%}&\textbf{95.2049}&93.9271&94.5879&91.5510&\multicolumn{1}{c|}{94.4082}&
			\multicolumn{1}{|c|}{\multirowcell{3}{$Min$\\$\downarrow$\\$Maj$}}&\multicolumn{1}{c|}{10\%}&\textbf{95.3747}&94.4065&94.4006&93.1466&\multicolumn{1}{c|}{95.2802}\\ 
			\multicolumn{1}{|c|}{\multirow{3}{*}{}}&\multicolumn{1}{c|}{20\%}&89.8330&90.1335&\textbf{93.3356}&90.1837&\multicolumn{1}{c|}{93.1020}&
			\multicolumn{1}{|c|}{\multirow{3}{*}{}}&\multicolumn{1}{c|}{20\%}&93.6558&93.2494&93.2268&92.7974&\multicolumn{1}{c|}{\textbf{94.2328}}\\ 
			\multicolumn{1}{|c|}{\multirow{3}{*}{}}&\multicolumn{1}{c|}{30\%}&84.8000&85.1669&87.0350&\textbf{89.3265}&\multicolumn{1}{c|}{88.2653}&
			\multicolumn{1}{|c|}{\multirow{3}{*}{}}&\multicolumn{1}{c|}{30\%}&90.9957&90.2254&\textbf{92.3607}&86.1983&\multicolumn{1}{c|}{92.0862}\\
			\hline
			\multicolumn{2}{|c|}{Dataset 9}&\multicolumn{1}{c|}{CE}&\multicolumn{1}{c|}{MSE}&\multicolumn{1}{c|}{C-Loss}&\multicolumn{1}{c|}{QMEE}&\multicolumn{1}{c|}{RMEE}&
			\multicolumn{2}{|c|}{Dataset 10}&\multicolumn{1}{c|}{CE}&\multicolumn{1}{c|}{MSE}&\multicolumn{1}{c|}{C-Loss}&\multicolumn{1}{c|}{QMEE}&\multicolumn{1}{c|}{RMEE}\\ 
			\hline
			\multicolumn{1}{|c|}{\multirowcell{3}{$Maj$\\$\downarrow$\\$Min$}}&\multicolumn{1}{c|}{10\%}&93.7470&92.4815&94.7562&90.8042&\multicolumn{1}{c|}{\textbf{95.3968}}&
			\multicolumn{1}{|c|}{\multirowcell{3}{$Maj$\\$\downarrow$\\$Min$}}&\multicolumn{1}{c|}{10\%}&\textbf{90.1181}&90.0219&90.0197&87.1096&\multicolumn{1}{c|}{90.0274}\\ 
			\multicolumn{1}{|c|}{\multirow{3}{*}{}}&\multicolumn{1}{c|}{20\%}&90.2963&90.3999&91.4588&87.8254&\multicolumn{1}{c|}{\textbf{92.5979}}&
			\multicolumn{1}{|c|}{\multirow{3}{*}{}}&\multicolumn{1}{c|}{20\%}&88.5257&88.5416&89.3293&85.8973&\multicolumn{1}{c|}{\textbf{90.3493}}\\ 
			\multicolumn{1}{|c|}{\multirow{3}{*}{}}&\multicolumn{1}{c|}{30\%}&84.8814&84.8500&90.5998&87.1270&\multicolumn{1}{c|}{\textbf{91.3228}}&
			\multicolumn{1}{|c|}{\multirow{3}{*}{}}&\multicolumn{1}{c|}{30\%}&85.6682&85.2683&85.3556&85.1781&\multicolumn{1}{c|}{\textbf{86.2945}}\\
			\hline
			\multicolumn{1}{|c|}{\multirowcell{3}{$Min$\\$\downarrow$\\$Maj$}}&\multicolumn{1}{c|}{10\%}&94.3642&\textbf{94.4588}&93.8574&91.3704&\multicolumn{1}{c|}{94.4392}&
			\multicolumn{1}{|c|}{\multirowcell{3}{$Min$\\$\downarrow$\\$Maj$}}&\multicolumn{1}{c|}{10\%}&\textbf{89.8244}&88.2781&87.9951&85.1712&\multicolumn{1}{c|}{88.2603}\\ 
			\multicolumn{1}{|c|}{\multirow{3}{*}{}}&\multicolumn{1}{c|}{20\%}&91.8626&\textbf{93.6889}&93.2479&90.7672&\multicolumn{1}{c|}{92.7090}&
			\multicolumn{1}{|c|}{\multirow{3}{*}{}}&\multicolumn{1}{c|}{20\%}&82.7101&83.3028&83.3895&84.8288&\multicolumn{1}{c|}{\textbf{84.9726}}\\ 
			\multicolumn{1}{|c|}{\multirow{3}{*}{}}&\multicolumn{1}{c|}{30\%}&88.9716&88.2154&89.0462&89.4127&\multicolumn{1}{c|}{\textbf{90.3085}}&
			\multicolumn{1}{|c|}{\multirow{3}{*}{}}&\multicolumn{1}{c|}{30\%}&79.3729&80.0598&80.1528&\textbf{82.4795}&\multicolumn{1}{c|}{80.5753}\\
			\hline
			\bottomrule
	\end{tabular}}
\end{table*}

In both TABLE \ref{tab_ben_att} and TABLE \ref{tab_ben_lab}, one can observe as before that RMEE achieves the highest accuracy under most circumstances. More detailed discussions about the experimental results will be proceeded in the next section.

\section{Discussion}
\label{sec7}
\subsection{RMEE vs QMEE}
This study aims to explore the implementation of MEE for robust classification. QMEE can be regarded as a simplified version of MEE, which utilizes the quantization technique. To this end, a straightforward idea is to apply the QMEE-based (or MEE-based) classifiers to contaminated datasets. Nevertheless, as one can see in experimental results, QMEE fails to achieve the excellent robustness as we expected. Particularly, in many cases of Fig. \ref{fig_attri_out_toy}, TABLE \ref{tab_ben_att}, and TABLE \ref{tab_ben_lab}, QMEE even achieves worse performance than conventional CE and MSE, which means robustness can not be realized by directly applying QMEE (or MEE) to classifiers.

RMEE can be regarded as a special case of QMEE with the predetermined codebook $C=(0,-1,1)$, which aims to drive the error PDF $f_E(e)$ towards the optimal one $\rho_E(e)$. Although QMEE has a more generalized form than RMEE, we would like to argue that it is the QMEE's larger degree of freedom that makes it inferior in robust classification. Proved by the extensive experimental results, RMEE achieves significantly better performance than QMEE in robust classification, which demonstrates superiority of the proposed restriction.

\subsection{RMEE vs C-Loss}
\label{rmee_closs}
C-Loss can be regarded as a special case of RMEE, when $\varPhi=(\varphi_0,\varphi_{-1},\varphi_{1})=(N,0,0)$, which aims to concentrate all errors around zero as possible. By contrast, RMEE permits some errors to be distributed at the worst cases, i.e. $e=\pm 1$. In this way, RMEE achieves better robustness than C-Loss in most cases, as proved by experimental results in Fig. \ref{fig_attri_out_toy}, Fig. \ref{fig_label_out_toy}, TABLE \ref{tab_ben_att}, and TABLE \ref{tab_ben_lab}.

On the other hand, considering the determination of $\varPhi$ for RMEE, as described in Subsection \ref{hyper}, RMEE actually uses C-Loss as initialization. The validity of such method to estimate the real outlier proportion needs to be further studied, which is beyond the scope of this paper. Moreover, except this method, how to obtain a more accurate estimation of the real outlier proportion without any prior information is a problem worth studying in the future.

\subsection{Extension to Other Classifiers}
As categorized in \cite{de2013minimum}, classifiers can be divided into regression-like and non-regression-like ones, in which prediction errors are of continuous and discrete values, respectively. For the regression-like classifiers, such as a wide variety of neural networks for classification, we argue that the proposed RMEE could be a promising alternative for those tasks prone to severe noises, since its effectiveness has been preliminarily verified on the ELM model in this paper.

On the other hand, the implementation of RMEE for non-regression-like classifiers needs further exploration. For example, in the decision trees and the $\{0,1\}$-label context, the prediction is discrete $0$ or $1$, and hence one obtains discrete error $e\in \{0,-1,1\}$, but not $e\in (-1,1)$ that belongs to a continuous interval as in this paper. Whether the proposed RMEE could achieve satisfactory performance for non-regression-like classifiers requires further studies.

\subsection{Extension to Multi-class Classification}
Considering the multi-class cases, each class has individual discriminant parameter $\{\omega_j\}_{j=1}^{\mathcal{Z}}$, where $\mathcal{Z}$ is the number of classes. The probability that \emph{i}th sample belongs to \emph{j}th class is calculated by \emph{softmax} as
\begin{equation}
y_{i}^{j}=P\left( t_i=j \right) =\frac{\exp \left( \omega _{j}'x_i \right)}{\sum_{j=1}^\mathcal{Z}{\exp \left( \omega _{j}'x_i \right)}} \,\, \left( j=1,...,\mathcal{Z} \right)
\end{equation}
In multi-class cases, \emph{one-hot} coding scheme is usually used for label denotation, e.g. $t_i=\left[ 1,0,...,0 \right] \in \mathbb{R}^\mathcal{Z}$ when the \emph{i}th sample is of the first class. Thus, one can similarly obtain multi-dimensional errors $\{e_i\}_{i=1}^{N} \in \mathbb{R}^\mathcal{Z}$ of continuous values by subtraction $e_i=t_i-p_i$, where $p_i=\left[ p_{i}^{1},p_{i}^{2},...,p_{i}^{\mathcal{Z}} \right]$. Extended from binary case, one can imagine that in multi-class cases errors are distributed on a high-dimensional cube ranging between $(-1,1)$. In this way, the implementation of RMEE for multi-class classification can refer to those studies that apply MEE to tasks with multi-dimensional errors, such as principal component analysis \cite{he2010principal}. Detailed implementation needs further argumentation.

\subsection{Topics for Future Studies}
In addition, we would like to state some topics for future studies of the proposed RMEE as follows. 

\noindent \textbf{1}: As stated in Subsection \ref{rmee_closs}, it is an interesting future work to obtain a more accurate estimation of the real outlier proportion without any prior information.

\noindent \textbf{2}: Comparing TABLE \ref{tab_ben_att} and TABLE \ref{tab_ben_lab}, one could find that the robustness of RMEE in label contamination is not as admirable as that in attribute contamination. Probably this is because that the target distribution $\rho_E(e)$ is not always appropriate for label contamination. The adverse samples caused by label contamination may not deviate significantly from the data cluster, and hence the corresponding errors are not large enough to $\pm1$. Therefore, different $\rho_E(e)$ for attribute and label contamination, respectively, could probably realize better robustness.

\section{Conclusion}
\label{sec8}
In conclusion, we explore the potential of MEE for robust classification against attribute and label outliers, proposing a novel variant by restricting MEE with a predetermined codebook. For the proposed RMEE, we discuss about its optimization and convergence analysis. By evaluating RMEE with logistic regression model and ELM model on toy datasets and benchmark datasets, respectively, we prove the encouraging capability of RMEE for robust classification. In addition, we provide some future topics for RMEE, which will be the crucial points for further improvements.

\ifCLASSOPTIONcaptionsoff
  \newpage
\fi

\bibliography{bibli}
\bibliographystyle{IEEEtran}
\end{document}